\documentclass[letterpaper]{article} 
\usepackage{aaai2026}  
\usepackage{times}  
\usepackage{helvet}  
\usepackage{courier}  
\usepackage[hyphens]{url}  
\usepackage{graphicx} 
\usepackage{natbib}  
\usepackage{caption} 
\usepackage{algorithm}
\usepackage{algorithmic}
\usepackage{booktabs} 
\usepackage{amsmath}
\usepackage{color} 
\usepackage{xcolor}
\usepackage{bm}
\usepackage{amsmath}
\usepackage{amssymb}
\usepackage{amsfonts}
\usepackage{graphicx}
\definecolor{purple}{RGB}{126,92,173}
\definecolor{blue}{RGB}{37,117,255}
\definecolor{green}{rgb}{0.0, 0.5, 0.0}
\definecolor{red}{rgb}{0.8, 0.0, 0.0}

\usepackage{multirow}
\usepackage{tabularx}
\usepackage{makecell}
\usepackage{bm}
\usepackage{newfloat}
\usepackage{listings}
\DeclareCaptionStyle{ruled}{labelfont=normalfont,labelsep=colon,strut=off} 
\floatstyle{ruled}
\newfloat{listing}{tb}{lst}{}
\floatname{listing}{Listing}
\title{Bias-Restrained Prefix Representation Finetuning for Mathematical Reasoning}
\author{
    Sirui Liang\textsuperscript{\rm 1,2,3}, Pengfei Cao\textsuperscript{\rm 1,2}\footnotemark[1], Jian Zhao\textsuperscript{\rm 3,4}, Cong Huang\textsuperscript{\rm 3,4}, Jun Zhao\textsuperscript{\rm 1,2}, Kang Liu\textsuperscript{\rm 1,2}\thanks{Corresponding author.}\\
}
\affiliations{
    \textsuperscript{\rm 1}The Key Laboratory of Cognition and Decision Intelligence for Complex Systems,\\ Institute of Automation, Chinese Academy of Sciences, Beijing, China\\
    \textsuperscript{\rm 2}School of Artificial Intelligence, University of Chinese Academy of Sciences, Beijing, China\\
    \textsuperscript{\rm 3}Zhongguancun Academy, Beijing, China\\
    \textsuperscript{\rm 4}Zhongguancun Institute of Artificial Intelligence, Beijing, China\\
     liangsirui2024@ia.ac.cn, \{pengfei.cao, jzhao, kliu\}@nlpr.ia.ac.cn, \{jianzhao, huangcong\}@zgci.ac.cn
}

\usepackage{bibentry}

\begin{document}
\maketitle
\begin{abstract}

Parameter-Efficient finetuning (PEFT) enhances model performance on downstream tasks by updating a minimal subset of parameters. Representation finetuning (ReFT) methods further improve efficiency by freezing model weights and optimizing internal representations with fewer parameters than PEFT, outperforming PEFT on several tasks. However, ReFT exhibits a significant performance decline on mathematical reasoning tasks. To address this problem, the paper demonstrates that ReFT's poor performance on mathematical tasks primarily stems from its struggle to generate effective reasoning prefixes during the early inference phase. Moreover, ReFT disturbs the numerical encoding and the error accumulats during the CoT stage. Based on these observations, this paper proposes \textbf{Bias-REstrained Prefix Representation FineTuning (BREP ReFT)}, which enhances ReFT's mathematical reasoning capability by truncating training data to optimize the generation of initial reasoning prefixes, intervening on the early inference stage to prevent error accumulation, and constraining the intervention vectors' magnitude to avoid disturbing numerical encoding. Extensive experiments across diverse model architectures demonstrate BREP's superior effectiveness, efficiency, and robust generalization capability, outperforming both standard ReFT and weight-based PEFT methods on the task of mathematical reasoning. The source code is available at https://github.com/LiangThree/BREP. 

\end{abstract}


\section{Introduction}
Large language models (LLMs) exhibit remarkable performance across a wide range of tasks \cite{achiam2023gpt, anil2023palm}. Task-specific finetuning further enhances LLMs' capabilities \cite{dai2015semi}. However, full parameter finetuning consumes substantial computational costs \cite{brown2020language, houlsby2019parameter}. Parameter-Efficient finetuing (PEFT) methods \cite{asai2022attempt, he2021towards}, such as Low-Rank Adaptation (LoRA) \cite{hu2022LoRA}, Prefix-tuning \cite{li2021prefix} and Adapter \cite{houlsby2019parameter}, mitigate this issue by updating a subset of full parameters, yet still require a significant number of trainable parameters \cite{ding2023sparse, ding2022delta, aghajanyan2020intrinsic}. A recently proposed finetuning paradigm, Representation Finetuning (ReFT), modifies the internal representations of the model without adjusting model parameters, and the model output could be intervened and modified targetly. In this way, greater efficiency could be achieved than PEFT \cite{li2023inference, wu2024reft, wu2024advancing}.


Actually, there are several adaptation paradigms for ReFT, including per-layer bias injection \cite{li2023inference}, scaling and biasing operations on layer-wise representations \cite{wu2024advancing}, and representation interventions within linear subspaces spanned by low-rank projection matrices \cite{wu2024reft}. This work focuses on the most generalizable and widely applicable ReFT approach: applying \textbf{learnable scaling and biasing operations} to the model's intermediate representations. Other ReFT methodologies can be conceptualized as specific instances or variations of this fundamental operation\cite{li2023inference, wu2024reft}.

ReFT demonstrates superior parameter efficiency and achieves better performance in commonsense reasoning, instruction-following \cite{ li2023inference, wang2025survey} but exhibits a significant performance gap in mathematical reasoning compared to LoRA, averaging 11.5\% in the GSM8K dataset. \citet{wu2024reft} indicated that ReFTs may have trouble in CoT reasoning than single-step commonsense reasoning tasks \cite{wei2022chain, cobbe2021training, hendrycks2021measuring}. Some representation engineering methods explored intervention techniques to enhance model reasoning ability \cite{tang2025unlocking, hojer2025improving}. However, the improvement achieved by these methods is small compared to the weight-based PEFT strategies.

\begin{figure*}[htbp]
\centering
\vspace{-10pt}
\includegraphics[width=2\columnwidth]{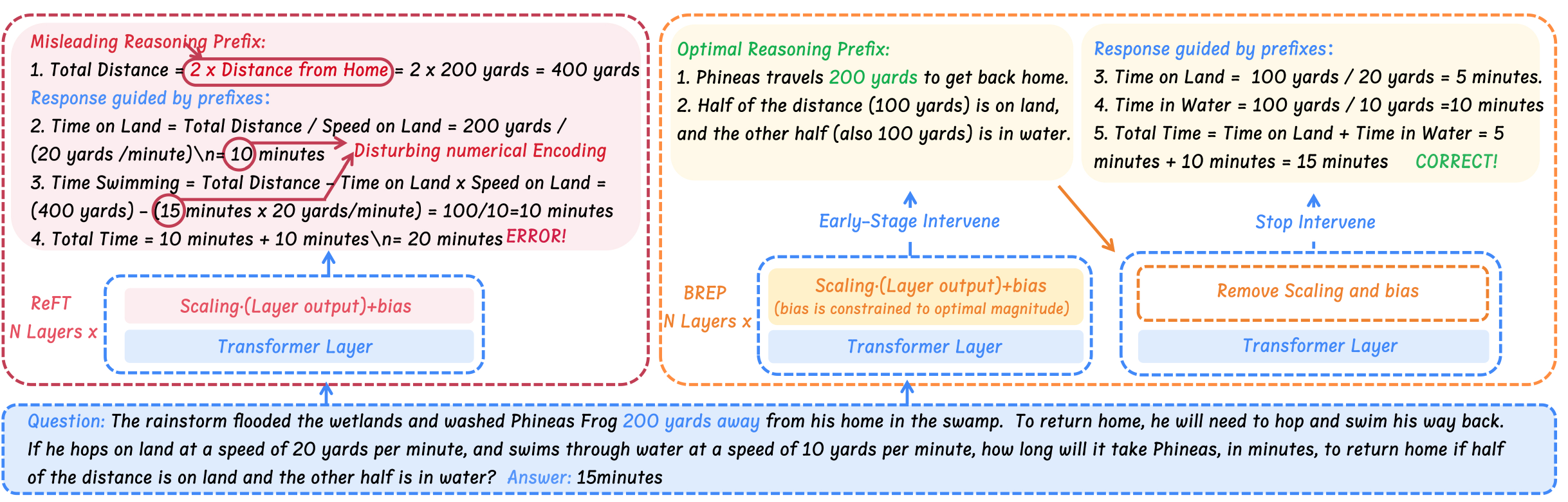}
\caption{The overview of ReFT and BREP. An example of misleading reasoning prefix and disturbing numerical encoding are shown in ReFT.
}
\vspace{-10pt}
\label{figure1}
\end{figure*}

Our diagnostic analysis reveals two primary factors for this failure (as shown in Figure \ref{figure1}): \textbf{1) Misleading reasoning prefixes}: ReFT struggles to generate effective initial reasoning steps (\textit{the first k tokens generated by model when answering questions}), which are critical for guiding the model towards a correct solution path \cite{ji2025first, qi2025shallow}. \textbf{2) Disturbing numerical encoding}: The intervention vectors used by ReFT can disrupt the model's internal representation of numbers, which is a fundamental component of mathematical reasoning. During the autoregressive generation of CoT, this error continues to accumulate as the output length increases, directly leading to the calculation error \cite{cao2024life, zhang2025crmr}.

Based on the aforementioned observations, this work proposes \textbf{Bias-Restrained Prefix  Representation Finetuning (BREP)}. It is a ReFT method that enhances mathematical reasoning performance through two components: \textbf{1) Prefix Training and Early-Stage Intervention}: Training data is truncated to focus on initial mathematical reasoning steps, learning high-quality prefix representations. During inference, applying intervention only to the first $k$ tokens, which maximizes the benefits of prefix training and prevents bias accumulation in CoT stage. \textbf{2) Bias Constraint Training}: Excessively large-magnitude intervene vectors damages the model's numerical encoding, while too short intervene vectors results  in ineffective utilization of ReFT's enhancements. This paper leverages Proportional-Integral-Derivative (PID) \cite{aastrom2006advanced} control during the train process, enabling intervene vectors constrained within an optimal magnitude. BREP significantly improves the performance of the model in mathematical reasoning tasks surpassing existing ReFT and PEFT methods while demonstrating robust generalization capabilities. The contributions of our work are summarized as follows:
\begin{itemize}
    \item Our analysis reveals that the primary causes of ReFT's underperformance in mathematical reasoning tasks are misleading initialization of reasoning prefix and excessively large-magnitude intervene vector disturbing numerical encoding.
    \item The paper proposes bias-restrained prefix ReFT method that enhances mathematical reasoning ability by optimizing the reasoning prefix and regulating intervention vector magnitudes to avoid numerical encoding error.
    \item A comprehensive experiment is conducted to demonstrate the effectiveness, efficiency, and robust generalization of our method.
\end{itemize}

\section{ReFT Definition}

In this section, a definition of ReFT is provided. For an input token sequence $\bm{x} = (x_1, \ldots, x_n)$ tokens are first embedded into vectors $\bm{h}^{0} = (\bm{h}_1^{0}, \ldots, \bm{h}_n^{0})$ where each $\bm{h}_i^{0} \in \bm{R}^d$. They are processed by $l$ transformer layers. Each layer $j$ ($1 \leq j \leq l$) computes:
\begin{equation}
\bm{h}^{j} = \bm{h}^{j-1} + \mathbf{A}^{j} + \mathbf{F}^{j}.
\label{eq:transformerlayer}
\end{equation}
with $\mathbf{A}^j$ and $\mathbf{F}^j$ being multi-head self-attention and feed-forward network outputs respectively.

A widely used ReFT adjustment \cite{wu2024advancing} method can be formalized as:
\begin{equation}
\bm{h}^{j} = \mathbf{W} \odot \left( \bm{h}^{j-1} + \mathbf{A}^{j} + \mathbf{F}^{j} \right) + \bm{b},
\label{eq:reft}
\end{equation}
where $\mathbf{W}, \bm{b} \in \bm{R}^d$ are learnable parameters, $\odot$ denotes element-wise multiplication (Hadamard product), and $\bm{h}^{j}$ is the ReFT modified representation. These vectors directly guide the model output and its magnitude and direction directly determine the effectiveness of the ReFT.

\section{Problem Diagnosis}

In this section, interpretable analyze experiments are conducted to reveal critical problems within ReFT in mathematical reasoning tasks. 

\subsection{Misleading Reasoning Prefixes}

\begin{figure}[t]
\centering
\vspace{-10pt}
\includegraphics[width=0.85\columnwidth]{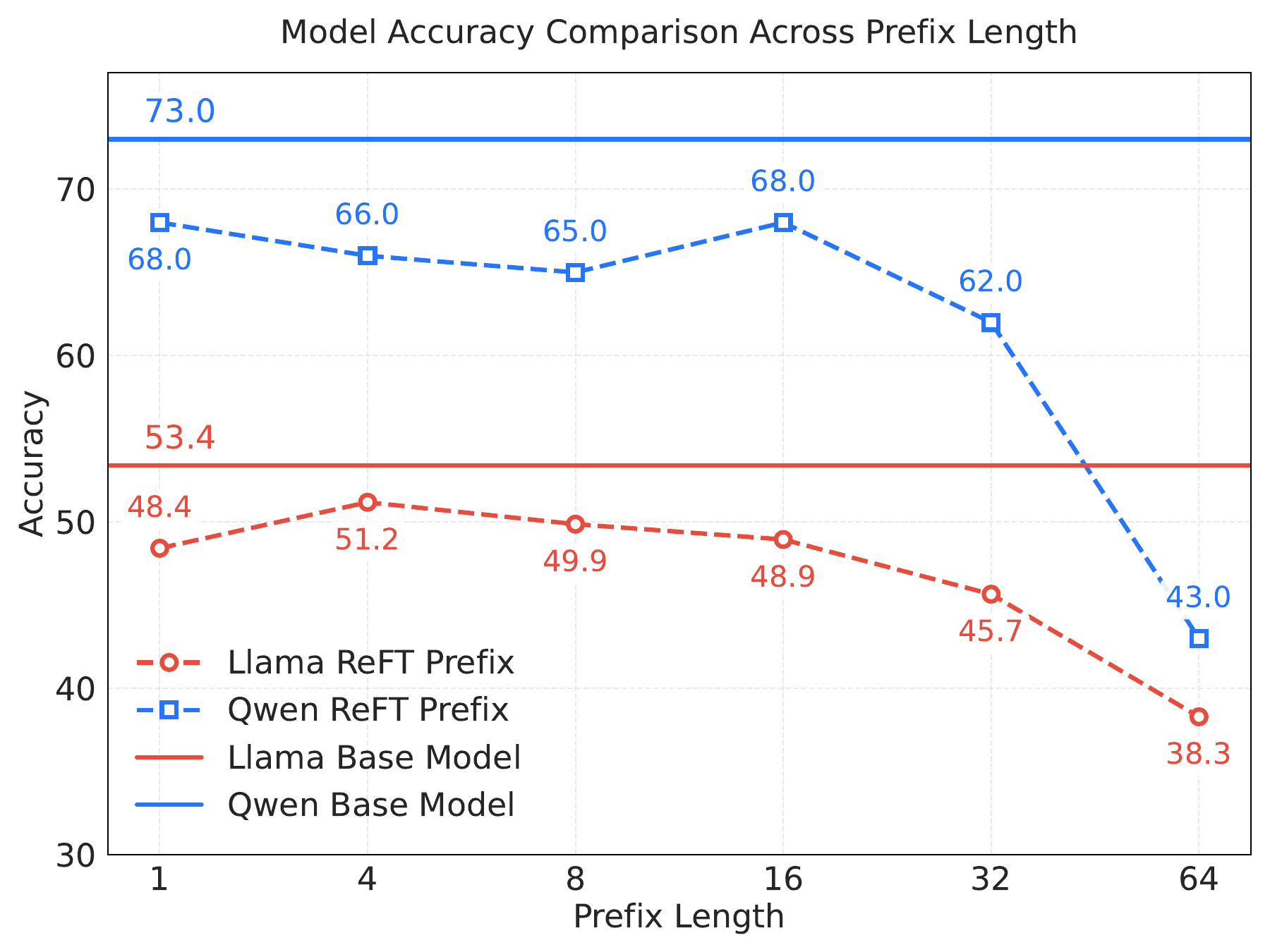}
\vspace{-5pt}
\caption{The impact of ReFT and Base prefix with various prefix length on mathematical reasoning performance.}
\vspace{-10pt}
\label{prefix_line}
\end{figure}

ReFT always fails to generate optimal reasoning steps during the early stage of mathematical problem solving. To validate this, prefixes of varying lengths were generated from a ReFT-finetuned model on 100 GSM8K \cite{cobbe2021training} cases. Under a temperature setting of 0.6, the base model continues to generate 10 responses along the given reasoning prefix. The answer results of the base model without prefix are also provided for comparison. As shown in Figure \ref{prefix_line}, shorter ReFT prefixes marginally underperform the base model without prefix, and the performance gap becomes more pronounced with longer ReFT prefixes. These observations suggest a suboptimal effectiveness of ReFT in early stage of mathematical reasoning. It is necessary to enhance the capability of ReFT in the reasoning build-stage and resolve the critical performance decline caused by longer representation intervention \cite{xu2025prompting}.

\subsection{Disturbing Numerical Encoding}

\paragraph{Theoretical Analysis} Current research provides evidence that LLMs encode the value of number linearly \cite{zhu2025language}. Formally, given a set of hidden states $\mathcal{H} = \{\bm{h}_1, \bm{h}_2, \ldots, \bm{h}_n\}$ and their corresponding original numbers $\mathcal{X} = \{x_1, x_2, \ldots, x_n\}$, a linear regressor $P$ can be trained to predict the number encoded in the internal representation of LLM:
\begin{equation}
P = \mathbf{N}\bm{h} + \bm{d},
\label{eq:numprob}
\end{equation}
where $\mathbf{N} \in \bm{R}^{d}$ and $\bm{d}$ are the weight parameters of $P$.

It is reasonable to hypothesize that, when ReFT modifies the hidden states, it may have an impact on digital encoding. Considering that ReFT induces minimal adjustments to weight parameter $\mathbf{W}$ during mathematical task training, this work focuses on analyzing the influence of bias parameter $\bm{b}$. Define a linear probe parameterized by $(\mathbf{N}, \bm{d})$. When ReFT introduces a bias intervention $\bm{\alpha}$ at a particular layer during inference, the perturbed numerical encoding becomes:
\begin{equation}
\hat{P} = \mathbf{N}(\bm{h} + \bm{\alpha}) + \bm{d} = \mathbf{N}\bm{h} + \bm{d} + \mathbf{N}\bm{\alpha} = P + \mathbf{N}\bm{\alpha},
\end{equation}
where $P$ denotes the original probe prediction.  When $\bm{\alpha}$ only affects the number in the direction of the digital encoding, $\bm{\alpha}$ is mapped to $\mathbf{N}$:
\begin{equation}
\hat{P} \approx P + \mathbf{N}(\mathbf{N} \cdot \bm{\alpha}) = P + c\|\mathbf{N}\|^2,
\end{equation}
where $c$ denotes the projection coefficient of $\alpha$ onto $N$. Generalizing to a complete forward pass through $L$ layers:
\begin{equation}
\hat{P}_{cum} = P + \left( \sum_{i=1}^{L} c_i \right) \|\mathbf{N}\|^2.
\end{equation}
As token generated during autoregressive inference, \textbf{this systematic error induced by ReFT progressively accumulates during CoT outputs}.

\begin{figure}[t]
\centering
\vspace{-10pt}
\includegraphics[width=1.0\columnwidth]{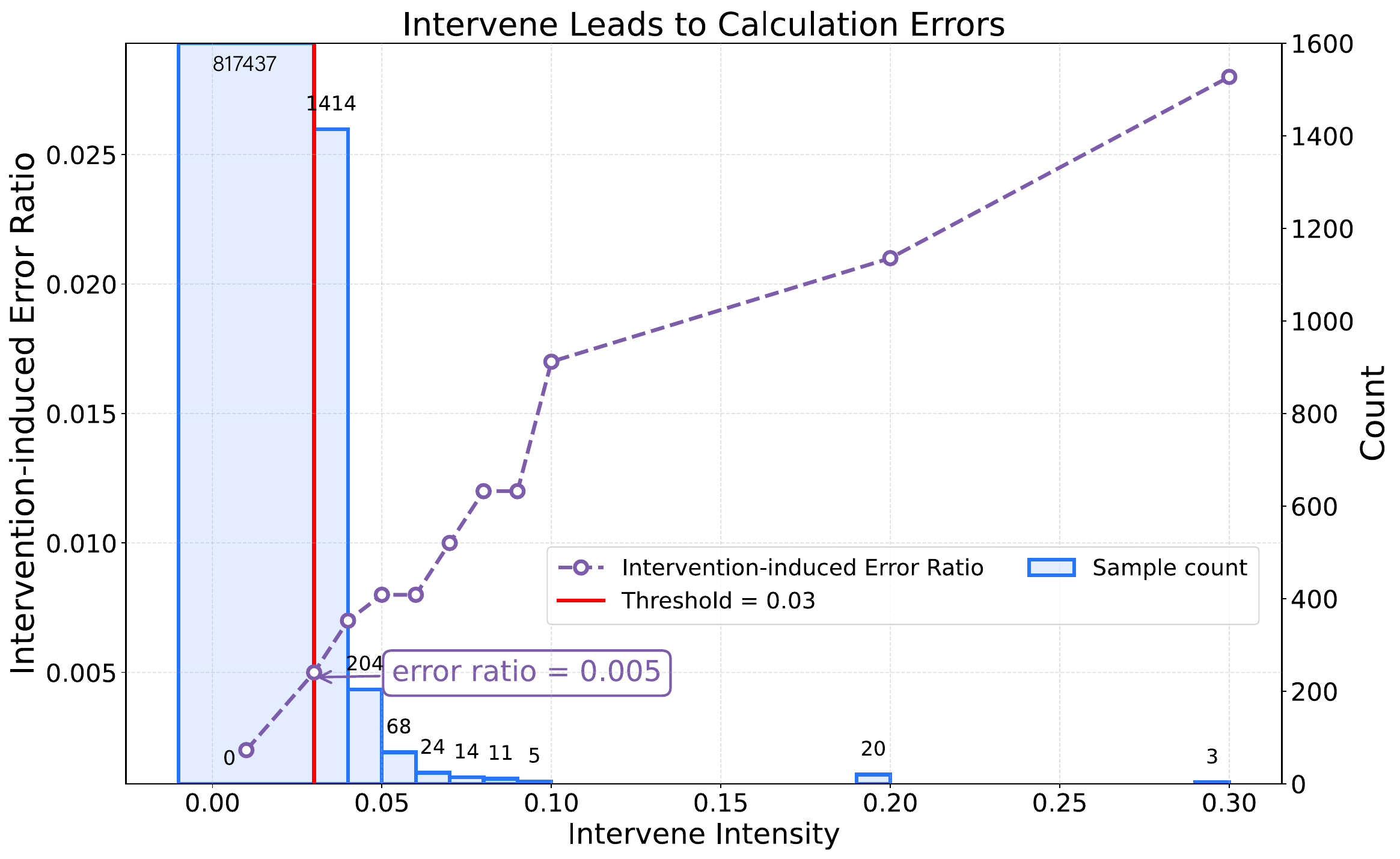} 
\caption{Effect of interventions on numerical encoding. X-axis: The intensity of intervention along the direction of numerical coding. \textcolor{purple}{Line plot (left y-axis)}: Error probability of four-digit addition under interventions. \textcolor{blue}{Bar plot (right y-axis)}: Intervention intensity of ReFT projected onto the number encoding direction.}
\vspace{-12pt}
\label{intervene}
\end{figure}

\paragraph{Empirical Validation} To validate our analysis, this paper first quantifies the intervention intensity threshold that disrupts the computational capacity of the model. During four-digit addition computations, \textbf{positive} interventions of magnitude $\delta$ along the numerical encoding direction ($\hat{n} = \frac{N}{\|N\|}$) are applied to the model. An intervention-induced error is defined as the calculation result after intervention is changed:
\begin{equation}
\text{Model}_{\text{positive-intervention}}(\delta) > \text{Model}_{\text{pre-intervention}},
\end{equation}
$\text{Model}_{\text{positive-intervention}}$ adds positive intervene to the model which increases the calculation result while negative intervene decreases it (model calculates 1234+4321=5560 positive-intervention $>$ 5555 pre-intervention). As shown in Figure \ref{intervene} \textcolor{purple}{(purple curve)}, as the intensity of interference increases, the error rate increases. This work establishes $\delta=0.04$ as the critical threshold, interventions beyond this magnitude induce $>$5 computational errors per 1,000 arithmetic problems.

Subsequently, the intervention intensity of ReFT vectors in the numerical encoding direction is measured. The hidden state deviation $\alpha$ in equation (4) is projected before and after ReFT onto the numerical encoding direction $\hat{n}$ (this quantifies the equivalent intervention intensity along the numerical encoding axis for bias introduced by ReFT). As illustrated in Figure \ref{intervene} \textcolor{blue}{(blue bar)}, more than 1,000 ReFT intervene vectors exceed the critical threshold ($\delta_{\text{crit}} = 0.04$). This statistically significant deviation directly explains 21 identified errors related to numbers (consist of confusing and forgetting key data that account for 18.4\% of all errors), empirically confirming that ReFT perturbs numerical encoding. The detailed numerical probe training process and comprehensive error statistics are documented in the Appendix.

\vspace{-5pt}
\section{Methodology}

Based on the aforementioned observations, this paper proposes \textbf{Bias-REstrained Prefix Representation Finetuning (BREP ReFT)}, which comprises two core components: 1) Prefix Training and Early-Stage Intervention, 2) Bias Constraint Training.

\subsection{Prefix Training and Early-Stage Intervention}

\paragraph{Prefix Training} Considering that the reasoning supervision signal is concentrated in early tokens \cite{qi2025shallow} and ReFT models always generate misleading reasoning prefixes, BREP \textbf{truncates the dataset and only utilizes the first $\bm{k}$ tokens} of each response sequence for training. The per-token supervision signal between the prefix training and full training is:
\begin{equation}
R_t(i)= \frac{1}{l_p}logp_t(y_t|x,y_{1:t-1}) - \frac{1}{l_f}logp_t(y_t|x,y_{1:t-1}),
\end{equation}
where $p_{t}$ denotes the conditional probability distribution generated by the model at position $t$,  $x$ represents question, $y_t$ is the target token. $l_p$ denotes the truncation length and $l_f$ is the complete sequence length. It can be observed that prefix training can provide a stronger signal strength when $t \leq l_p< l_f$. 

\paragraph{Early-Stage Intervention} During inference, BREP \textbf{applies interventions only to the first $\bm{n}$ tokens} ($n \leq l_p$). The optimization target becomes the prefix cumulative reward:
\begin{equation}
R_{cum} = \sum_{t=1}^n R_t(i).
\end{equation}
This enables full use of the prefix training supervision signal and avoids the accumulation of bias caused by continuous intervention. Furthermore, the optimized gradient directionality is as follows:
\begin{equation}
\nabla_{\theta} R_{\textit{cum}}^{\textit{intervene}} \propto \sum_{t=1}^n \nabla_{\theta} log p(y_t \mid x, y_{1:t-1}),
\end{equation}
which formulates an ideal target gradient which is derived exclusively from the first $n$ tokens of the sequence, $\nabla_{\theta} R$ represents the direction of the cumulative gradient of the training.
\begin{equation}
\nabla_{\theta} R_{\textit{cum}}^{\textit{prefix}} \propto \nabla_{\theta} R_{\textit{cum}}^{\textit{intervene}} + \underbrace{
 \sum_{t=n+1}^{l_p} \alpha_t \nabla_{\theta} log p(y_t \mid \cdot)}_{\textbf{contamination\ signal}},
\end{equation}
\vspace{-5pt}
\begin{equation}
\nabla_{\theta} R_{\textit{cum}}^{\textit{full}} \propto \nabla_{\theta} R_{\textit{cum}}^{\textit{intervene}}  + \underbrace{
 \sum_{t=n+1}^{l_f} \alpha_t \nabla_{\theta} log p(y_t \mid \cdot)}_{\textbf{contamination\ signal}},
\end{equation}
describes the gradient for prefix training and full training. Prefix training focuses on optimization of initial reasoning, while full training suffers significant contamination beyond position \textit{n}. 

In general, prefix truncation brings a stronger supervised signal and a gradient direction that is focused on early-stage reasoning thoughts. Early-stage intervention makes full use of these signals while avoiding the accumulation of errors caused by intervention.

\subsection{Bias Constraint Training}

To constrain the magnitude of the learned intervene vector, this paper develops a generalized framework for precision-controlled bias magnitude during training. The total loss function $\mathcal{L}_{\text{total}}$ consists of two components: a cross-entropy loss term and an adaptive bias magnitude adjustment.

\subsubsection{Cross-Entropy Loss}
Given target sequence $Y = \{y_1, y_2, \dots, y_k\}$ containing the first $k$ response tokens, the cross-entropy loss at timestep $t$ is computed as:
\begin{equation}
\mathcal{L}_{\text{ce}} = -\frac{1}{k} \sum_{t=1}^{k}logp_t({y_t \mid x, y_{1:t-1}}),
\end{equation}
where $p_t^{y_t} \in [0,1]$ denotes the softmax-normalized probability of the target token $y_t$ at position $t$, $x$ represents question and $V$ represents the vocabulary size ($\sum_{i=1}^{V} p_t^{i} = 1$).
\subsubsection{Adaptive Bias Magnitude Adjustment}
\label{subsec:weight_adjustment}
To accelerate early-stage learning while avoiding excessive bias in learning, an adaptive weight $w(t)$ is regulated via Proportional-Integral-Derivative (PID) control \cite{aastrom2006advanced, tohma2025smartcontrol} which represents a foundational control algorithm in dynamical systems. The weight update mechanism follows:
\begin{equation}
b(t) = \frac{1}{L} \sum_{l=1}^{L} \| \mathbf{b}_l(t) \|_2,
\end{equation}
where $b(t)$ denotes the mean bias vector learned across all layers at the current training step.
\begin{equation}
e(t) = b_{\text{target}} - b(t),
\end{equation}
where $b_{\text{target}}$ is the target bias magnitude ($b_{\text{target}}$ \textit{will be analyzed in subsection Optimal Bias Magnitude}) and $e(t)$ is the distance from the target bias magnitude.
\begin{equation}
\Delta w(t) = K_p \cdot e(t) + K_i \cdot \int_{0}^{t} e(t)  dt + K_d \cdot \frac{de(t)}{dt},
\end{equation}
where $K_p$ provides instantaneous error correction, $K_i$ eliminates steady-state offset through accumulated error integration, and $K_d$ enhances stability ($K_p=1\times 10^{-1}, K_i=1\times 10^{-4}, K_d=1\times 10^{-2}$). Our method automatically constrain the bias magnitude through $\Delta w(t)$.
\begin{equation}
w(t+1) =  clip( w(t) \cdot \left(1 + \alpha \cdot \Delta w(t)\right), w_{min}, w_{max}),
\end{equation}
$\alpha$  is a smoothing factor to prevent excessive fluctuations in the loss ($\alpha = 5$) and $ w_{max}, w_{min}$ ensure stable update range($ w_{max}=1\times 10^{-1}, w_{min}=1\times 10^{-5}$). 

Bias constrained training enables precise optimization of bias magnitude acquisition, ensuring models converge to empirically validated optimal magnitude. The total loss is formulated as:
\begin{equation}
\mathcal{L}_{\text{total}} = w(t) \cdot \mathcal{L}_{\text{ce}}.
\end{equation}
This enables the ReFT intervention vector to learn an optimal magnitude, ensuring its effectiveness while preventing disturbing the numerical encoding.

\section{Experiments} \label{sec:experiments}

\begin{table*}[ht]
\vspace{-10pt}
\centering
\scalebox{0.95}{
\begin{tabular}{cclllllll}
\toprule
\multirow{2}{*}{\textbf{Model}} & \multirow{2}{*}{\textbf{Method}} & \multicolumn{4}{c}{\textbf{Math10K}} & \multicolumn{3}{c}{\textbf{PRM800K}} \\
\cmidrule(lr){3-6} \cmidrule(lr){7-9}
 & & \textbf{GSM8K} & \textbf{SVAMP} & \textbf{MathQA} & \textbf{Avg} & \textbf{MATH500} & \textbf{AMC23} & \textbf{Avg} \\
\midrule
\multirow{5}{*}{\makecell{\textbf{Llama3-8B} \\ \textbf{-Instruct}}}
 & \textbf{Base} & 80.0 & 88.9 & \textbf{55.0} & 74.6 & 40.4 & \textbf{57.5} & \textbf{49.0} \\
 & \textbf{LoRA} & 81.1\,\textcolor{green}{\scriptsize{↑1.1}} & \textbf{90.0}\,\textcolor{green}{\scriptsize{↑1.1}} & 54.0\,\textcolor{red}{\scriptsize{↓1.0}} & 75.0\,\textcolor{green}{\scriptsize{↑0.4}} & 39.3\,\textcolor{red}{\scriptsize{↓1.1}} & 53.8\,\textcolor{red}{\scriptsize{↓3.7}} & 46.6\,\textcolor{red}{\scriptsize{↓2.4}} \\
 & \textbf{RED} & 73.8\,\textcolor{red}{\scriptsize{↓6.2}} & 88.9\,\textcolor{blue}{$=$} & 51.3\,\textcolor{red}{\scriptsize{↓3.7}} & 71.3\,\textcolor{red}{\scriptsize{↓3.3}} & 41.5\,\textcolor{green}{\scriptsize{↑1.1}} & 56.4\,\textcolor{red}{\scriptsize{↓1.1}} & \textbf{49.0}\,\textcolor{blue}{$=$} \\
 & \textbf{LoReFT} & 78.8\,\textcolor{red}{\scriptsize{↓1.2}} & 80.7\,\textcolor{red}{\scriptsize{↓8.2}} & 44.7\,\textcolor{red}{\scriptsize{↓10.3}} & 68.1\,\textcolor{red}{\scriptsize{↓6.5}} & 37.0\,\textcolor{red}{\scriptsize{↓3.4}} & 35.0\,\textcolor{red}{\scriptsize{↓22.5}} & 36.0\,\textcolor{red}{\scriptsize{↓13.0}} \\
 \cmidrule(lr){2-9}
 & \textbf{BREP\scriptsize{(ours)}} & \textbf{82.8}\,\textcolor{green}{\scriptsize{↑2.8}} & 89.5\,\textcolor{green}{\scriptsize{↑0.6}} & 54.3\,\textcolor{red}{\scriptsize{↓0.7}} & \textbf{75.5}\,\textcolor{green}{\scriptsize{↑0.9}} & \textbf{42.8}\,\textcolor{green}{\scriptsize{↑2.4}} & 52.5\,\textcolor{red}{\scriptsize{↓5.0}} & 47.7\,\textcolor{red}{\scriptsize{↓1.3}} \\
\midrule
\multirow{5}{*}{\makecell{\textbf{Llama3.1-8B} \\ \textbf{-Instruct}}}
 & \textbf{Base} & 87.3 & 94.4 & 72.8 & 84.8 & 59.8 & \textbf{77.5} & 68.7 \\
 & \textbf{LoRA} & 87.9\,\textcolor{green}{\scriptsize{↑0.6}} & 94.8\,\textcolor{green}{\scriptsize{↑0.4}} & 72.9\,\textcolor{green}{\scriptsize{↑0.1}} & 85.2\,\textcolor{green}{\scriptsize{↑0.4}} & 43.7\,\textcolor{red}{\scriptsize{↓16.1}} & 61.5\,\textcolor{red}{\scriptsize{↓16.0}} & 52.6\,\textcolor{red}{\scriptsize{↓16.1}} \\
 & \textbf{RED} & 82.9\,\textcolor{red}{\scriptsize{↓4.4}} & 91.1\,\textcolor{red}{\scriptsize{↓3.3}} & 67.5\,\textcolor{red}{\scriptsize{↓5.3}} & 80.5\,\textcolor{red}{\scriptsize{↓4.3}} & 58.1\,\textcolor{red}{\scriptsize{↓1.7}} & 70.0\,\textcolor{red}{\scriptsize{↓7.5}} & 64.1\,\textcolor{red}{\scriptsize{↓4.6}} \\
 & \textbf{LoReFT} & 84.1\,\textcolor{red}{\scriptsize{↓3.2}} & 89.1\,\textcolor{red}{\scriptsize{↓5.3}} & 57.0\,\textcolor{red}{\scriptsize{↓15.8}} & 76.7\,\textcolor{red}{\scriptsize{↓8.1}} & 42.6\,\textcolor{red}{\scriptsize{↓17.2}} & 42.5\,\textcolor{red}{\scriptsize{↓35.0}} & 42.6\,\textcolor{red}{\scriptsize{↓26.1}} \\
 \cmidrule(lr){2-9}
 & \textbf{BREP\scriptsize{(ours)}} & \textbf{91.6}\,\textcolor{green}{\scriptsize{↑4.3}} & \textbf{96.8}\,\textcolor{green}{\scriptsize{↑2.4}} & \textbf{73.3}\,\textcolor{green}{\scriptsize{↑0.5}} & \textbf{87.2}\,\textcolor{green}{\scriptsize{↑2.4}} & \textbf{63.5}\,\textcolor{green}{\scriptsize{↑3.7}} & \textbf{77.5}\,\textcolor{blue}{$=$} & \textbf{70.5}\,\textcolor{green}{\scriptsize{↑1.8}} \\
\midrule
\multirow{5}{*}{\makecell{\textbf{Qwen2.5-Math} \\ \textbf{-7B-Instruct}}}
 & \textbf{Base} & 95.4 & 96.1 & 82.7 & 91.4 & 81.2 & 60.0 & 70.6 \\
 & \textbf{LoRA} & 96.4\,\textcolor{green}{\scriptsize{↑1.0}} & \textbf{96.5}\,\textcolor{green}{\scriptsize{↑0.4}} & 83.9\,\textcolor{green}{\scriptsize{↑1.2}} & 92.3\,\textcolor{green}{\scriptsize{↑0.9}} & 80.8\,\textcolor{red}{\scriptsize{↓0.4}} & \textbf{74.4}\,\textcolor{green}{\scriptsize{↑14.4}} & 77.6\,\textcolor{green}{\scriptsize{↑7.0}} \\
 & \textbf{RED} & 95.7\,\textcolor{green}{\scriptsize{↑0.3}} & 95.9\,\textcolor{red}{\scriptsize{↓0.2}} & 81.7\,\textcolor{red}{\scriptsize{↓1.0}} & 91.1\,\textcolor{red}{\scriptsize{↓0.3}} & 78.2\,\textcolor{red}{\scriptsize{↓3.0}} & 66.7\,\textcolor{green}{\scriptsize{↑6.7}} & 72.5\,\textcolor{green}{\scriptsize{↑1.9}} \\
 & \textbf{LoReFT} & 93.6\,\textcolor{red}{\scriptsize{↓1.8}} & 95.3\,\textcolor{red}{\scriptsize{↓0.8}} & 74.5\,\textcolor{red}{\scriptsize{↓8.2}} & 87.8\,\textcolor{red}{\scriptsize{↓3.6}} & 78.8\,\textcolor{red}{\scriptsize{↓2.4}} & 72.5\,\textcolor{green}{\scriptsize{↑12.5}} & 75.7\,\textcolor{green}{\scriptsize{↑5.1}} \\
 \cmidrule(lr){2-9}
 & \textbf{BREP\scriptsize{(ours)}} & \textbf{96.9}\,\textcolor{green}{\scriptsize{↑1.5}} & 96.2\,\textcolor{green}{\scriptsize{↑0.1}} & \textbf{84.3}\,\textcolor{green}{\scriptsize{↑1.6}} & \textbf{92.5}\,\textcolor{green}{\scriptsize{↑1.1}} & \textbf{82.0}\,\textcolor{green}{\scriptsize{↑0.8}} & \textbf{74.4}\,\textcolor{green}{\scriptsize{↑14.4}} & \textbf{78.2}\,\textcolor{green}{\scriptsize{↑7.6}} \\
\midrule
\multirow{5}{*}{\textbf{Qwen3-8B}} 
 & \textbf{Base} & 95.1 & 96.7 & \textbf{86.5} & 92.8 & 82.0 & 85.0 & 83.5 \\
 & \textbf{LoRA} & 95.1\,\textcolor{blue}{$=$} & 96.8\,\textcolor{green}{\scriptsize{↑0.1}} & 86.2\,\textcolor{red}{\scriptsize{↓0.3}} & 92.7\,\textcolor{red}{\scriptsize{↓0.1}} & 81.8\,\textcolor{red}{\scriptsize{↓0.2}} & \textbf{87.5}\,\textcolor{green}{\scriptsize{↑2.5}} & 84.7\,\textcolor{green}{\scriptsize{↑1.2}} \\
 & \textbf{RED} & 87.9\,\textcolor{red}{\scriptsize{↓7.2}} & 91.8\,\textcolor{red}{\scriptsize{↓4.9}} & 77.3\,\textcolor{red}{\scriptsize{↓9.2}} & 85.7\,\textcolor{red}{\scriptsize{↓7.1}} & 54.2\,\textcolor{red}{\scriptsize{↓27.8}} & 35.0\,\textcolor{red}{\scriptsize{↓50.0}} & 44.6\,\textcolor{red}{\scriptsize{↓38.9}} \\
 & \textbf{LoReFT} & 87.1\,\textcolor{red}{\scriptsize{↓8.0}} & 96.3\,\textcolor{red}{\scriptsize{↓0.4}} & 72.8\,\textcolor{red}{\scriptsize{↓13.7}} & 85.4\,\textcolor{red}{\scriptsize{↓7.4}} & 72.4\,\textcolor{red}{\scriptsize{↓9.6}} & 80.0\,\textcolor{red}{\scriptsize{↓5.0}} & 76.2\,\textcolor{red}{\scriptsize{↓7.3}} \\
  \cmidrule(lr){2-9}
 & \textbf{BREP\scriptsize{(ours)}} & \textbf{95.3}\,\textcolor{green}{\scriptsize{↑0.2}} & \textbf{97.4}\,\textcolor{green}{\scriptsize{↑0.7}} & 86.3\,\textcolor{red}{\scriptsize{↓0.2}} & \textbf{93.0}\,\textcolor{green}{\scriptsize{↑0.2}} & \textbf{82.6}\,\textcolor{green}{\scriptsize{↑0.6}} & \textbf{87.5}\,\textcolor{green}{\scriptsize{↑2.5}} & \textbf{85.1}\,\textcolor{green}{\scriptsize{↑1.6}} \\
\midrule
\multirow{5}{*}{\textbf{Qwen3-14B}} 
 & \textbf{Base} & 96.5 & 96.3 & 87.4 & 93.4 & 84.2 & 82.5 & 83.4 \\
 & \textbf{LoRA} & 96.6\,\textcolor{green}{\scriptsize{↑0.1}} & 96.1\,\textcolor{red}{\scriptsize{↓0.2}} & 87.1\,\textcolor{red}{\scriptsize{↓0.3}} & 93.3\,\textcolor{red}{\scriptsize{↓0.1}} & 84.4\,\textcolor{green}{\scriptsize{↑0.2}} & 70.0\,\textcolor{red}{\scriptsize{↓12.5}} & 77.2\,\textcolor{red}{\scriptsize{↓6.2}} \\
 & \textbf{RED} & 93.5\,\textcolor{red}{\scriptsize{↓3.0}} & 95.8\,\textcolor{red}{\scriptsize{↓0.5}} & 82.9\,\textcolor{red}{\scriptsize{↓4.5}} & 90.7\,\textcolor{red}{\scriptsize{↓2.7}} & 63.2\,\textcolor{red}{\scriptsize{↓21.0}} & 57.5\,\textcolor{red}{\scriptsize{↓25.0}} & 60.4\,\textcolor{red}{\scriptsize{↓23.0}} \\
 & \textbf{LoReFT} & 88.5\,\textcolor{red}{\scriptsize{↓8.0}} & \textbf{96.8}\,\textcolor{green}{\scriptsize{↑0.5}} & 74.2\,\textcolor{red}{\scriptsize{↓13.2}} & 86.5\,\textcolor{red}{\scriptsize{↓6.9}} & 73.0\,\textcolor{red}{\scriptsize{↓11.2}} & 77.5\,\textcolor{red}{\scriptsize{↓5.0}} & 75.3\,\textcolor{red}{\scriptsize{↓8.1}} \\
 \cmidrule(lr){2-9}
 & \textbf{BREP\scriptsize{(ours)}} & \textbf{96.8}\,\textcolor{green}{\scriptsize{↑0.3}} & 96.2\,\textcolor{red}{\scriptsize{↓0.1}} & \textbf{87.6}\,\textcolor{green}{\scriptsize{↑0.2}} & \textbf{93.5}\,\textcolor{green}{\scriptsize{↑0.1}} & \textbf{84.6}\,\textcolor{green}{\scriptsize{↑0.4}} & \textbf{90.0}\,\textcolor{green}{\scriptsize{↑7.5}} & \textbf{87.3}\,\textcolor{green}{\scriptsize{↑3.9}} \\
\bottomrule
\end{tabular}
}
\caption{Performance comparison of Base (the original model without finetuning), RED \cite{wu2024advancing}, LoRA \cite{hu2022LoRA}, LoReFT \cite{wu2024reft} and BREP{\scriptsize{(ours)}}. Up arrows \textcolor{green}{(↑)} indicate improvement over base model, down arrows \textcolor{red}{(↓)} indicate decrease. Best method in each group marked in \textbf{bold}.}
\vspace{-10pt}
\label{tab:main_table}
\end{table*}
\subsection{Experimental Setup}

\paragraph{Datasets.} The experiments utilize two series of open-source LLMs (Llama \cite{grattafiori2024llama} and Qwen \cite{yang2025qwen3}) on mathematical reasoning tasks. Simple mathematical reasoning tasks are finetuned on a combination of arithmetic reasoning datasets MATH10K \cite{hu2023llm} and tested on standard benchmarks including GSM8K \cite{cobbe2021training}, SVAMP \cite{patel2021nlp} and MATHQA \cite{amini2019mathqa}. Complex mathematical reasoning tasks are finetuned on PRM800K \cite{lightman2023let} which includes step-level human feedback labels and test on competitive benchmarks (MATH500 \cite{hendrycks2021measuring} and AMC).

\paragraph{Baselines.} Different finetuning methods are compared, such as the ReFT method (RED \cite{wu2024advancing} and LoReFT \cite{wu2024reft}), weight-based PEFT methods (LoRA \cite{hu2022LoRA}), with Zero-shot CoT.

\paragraph{Implementation Details.} Our experiments use two series open-source LLMs: Llama3-8B-Instruct, Llama3.1-8B-Instruct \cite{grattafiori2024llama}, Qwen2.5-Math-7B-Instruct \cite{bai2023qwen}, Qwen3-8B \cite{yang2025qwen3} and Qwen3-14B. Each method was fine-tuned using 5,000 data. All of our experiments are run with a single Nvidia A100 80G GPU. More experimental details are provided in Appendix.

\subsection{Experimental Results}

To verify the effectiveness of our method in mathematical reasoning tasks, we compare it  with the ReFT and weight-based PEFT methods.
As shown in Table \ref{tab:main_table}. Vanilla ReFT methods (\textit{e.g.}, RED) and its variant LoReFT achieve performance marginally below the base model on mathematical reasoning tasks, while LoRA finetuning improves accuracy on most benchmarks. In contrast, BREP substantially enhances mathematical reasoning in nearly all models and test sets. Crucially, comparative analysis against weight-based PEFT methods reveals BREP's robustness: although LoRA achieves competitive results on simpler math reasoning tasks, it exhibits unstable performance on complex benchmarks (\textit{e.g.}, MATH500). BREP maintains strong performance across difficulty levels, demonstrating notable gains in challenging problems. This indicates that BREP's constraint representation learning enables stronger generalization for multi-step mathematical reasoning.

\subsection{Ablation Study}

\begin{figure*}[t]
\centering
\vspace{-15pt}
\includegraphics[width=2.2\columnwidth]{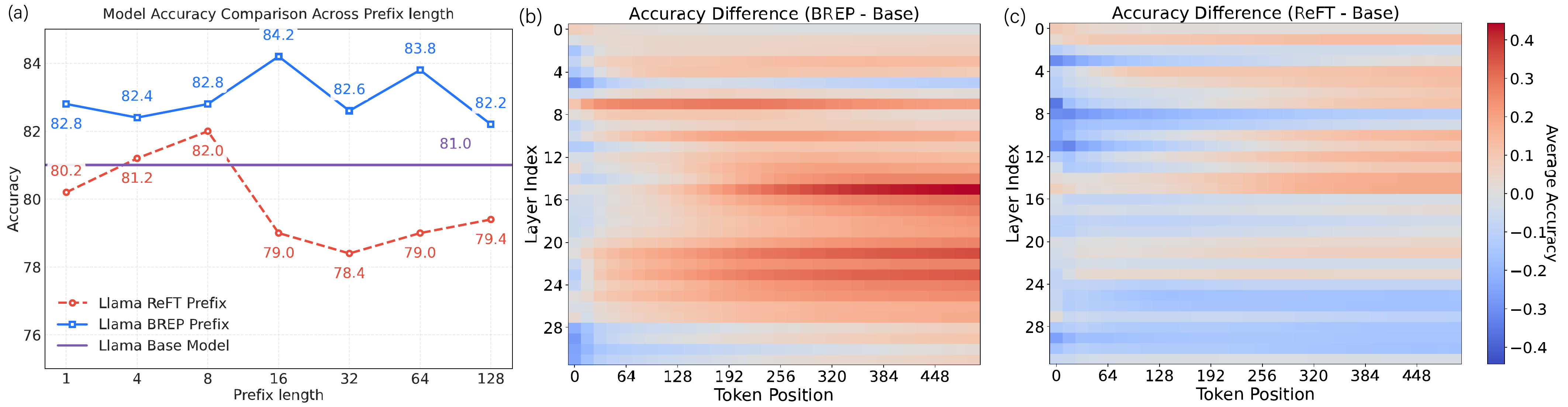}
\caption{The analysis of BREP. (a) Comparison of mathematical reasoning effectiveness guided by BREP prefixes and ReFT prefixes. (b) Numerical faithfulness performance gap between BREP and Base model. (c) Numerical faithfulness performance gap between ReFT and Base model. (\textcolor{red}{Red} indicates faithfulness improvement, \textcolor{blue}{blue} indicates faithfulness degradation.) }
\vspace{-10pt}
\label{sample_prefix}
\end{figure*}

\begin{table}[t]
\centering
\begin{tabular}{c l l l}
\toprule
\textbf{Model} & \textbf{Ablation} & \textbf{GSM8K} & \textbf{MATH500} \\
\midrule
\multirow{5}{*}{\makecell{\textbf{Llama3-8B}\\\textbf{-Instruct}}} 
& \textbf{BREP} & 82.8 & 42.7 \\
& \textbf{w/o PT} & 81.0\,\textcolor{red}{\scriptsize{↓1.8}} & 40.2\,\textcolor{red}{\scriptsize{↓2.5}} \\
& \textbf{w/o BCT} & 80.0\,\textcolor{red}{\scriptsize{↓2.8}} & 39.4\,\textcolor{red}{\scriptsize{↓3.3}} \\
& \textbf{w/o ESI} & 80.4\,\textcolor{red}{\scriptsize{↓2.4}} & 37.6\,\textcolor{red}{\scriptsize{↓5.1}} \\
\cmidrule(lr){1-4}
\multirow{5}{*}{\textbf{Qwen3-8B}} 
& \textbf{BREP} & 96.9 & 82.0 \\
& \textbf{w/o PT} & 95.5\,\textcolor{red}{\scriptsize{↓1.4}} & 79.4\,\textcolor{red}{\scriptsize{↓2.6}} \\
& \textbf{w/o BCT} & 94.9\,\textcolor{red}{\scriptsize{↓2.0}} & 79.8\,\textcolor{red}{\scriptsize{↓2.2}} \\
& \textbf{w/o ESI} & 95.1\,\textcolor{red}{\scriptsize{↓1.8}} & 81.6\,\textcolor{red}{\scriptsize{↓0.4}} \\
\bottomrule
\end{tabular}
\caption{Ablation study of BREP, “w/o” denotes the removal of subcomponent. PT: Prefix Truncation, BCT: Bias Constraint Training, ESI: Early-Stage Intervention.}
\vspace{-10pt}
\label{tab:ablation}
\end{table}

Ablation studies are conducted by systematically removing individual modules. Table \ref{tab:ablation} demonstrates the critical contributions of three core components—Prefix Truncation (PT), Bias Constraint Training (BCT), and Early-Stage Intervention (ESI). The most pronounced decline observed when removing Bias Constraint Training, particularly for complex reasoning, validating its necessity in ensure accurate calculation through intervention vector regulation. Early-Stage Intervention removal harms performance similarly, this aligns with our hypothesis that interventions within longer reasoning chains enlarge error accumulation. Prefix Truncation's absence yields smaller declines. Notably, performance gaps widen with reasoning complexity: average drops on MATH500 (↓2.68\%) exceed those on GSM8K (↓2.03\%), underscoring BREP's enhanced efficacy for multi-step long CoT problems. 
These results establish the synergistic necessity of all components for BREP.


\subsection{BREP Exhibits Superior Reasoning Prefix}

To verify whether BREP effectively addresses ReFT's poor performance in generating reasoning prefixes during the early stage (analyzed in \textit{Misleading Reasoning Prefixes}), 500 problems were randomly selected from GSM8K and used both BREP and ReFT models to generate prefixes of varying lengths. The base model then continued answering based on the given problems and prefixes. As shown in Figure \ref{sample_prefix}(a), BREP consistently outperforms ReFT by +0.8–5.2\% accuracy across different prefix lengths, with its advantage becoming more pronounced as the prefix length increases. This demonstrates that BREP generates significantly higher-quality initial reasoning paths. In complex, multi-step reasoning scenarios, BREP can consistently enlarge the optimaztion gap with longer prefixes.

\subsection{BREP Encodes Numbers Faithfully} 

This section verifies whether BREP effectively mitigates the issue of ReFT in disturbing numerical encoding (analyzed in \textit{Deviation accumulation in CoT}). Probes are trained to detect digital faithfulness within the model \cite{li2023inference, belinkov2022probing}, the detailed training data construction method and probe accuracy are introduced in \textit{Probe Details} in Appendix. Figure \ref{sample_prefix}(b) (\textit{BREP-Base}) shows the accuracy gap between BREP and the Base model across all layers and token positions. Conversely, Figure \ref{sample_prefix}(c) (\textit{ReFT-Base}) illustrates the accuracy gap between ReFT and the Base model. It can be observed visually that BREP exhibited significantly higher numerical faithfulness compared to the Base model, while ReFT performed below the Base model's faithfulness level. Detailed data statistics are available in Appendix. This demonstrates the efficacy of BREP in resolving the disturbation of ReFT on numerical encoding.

\begin{figure}[tbp]
\centering
\includegraphics[width=0.9\columnwidth]{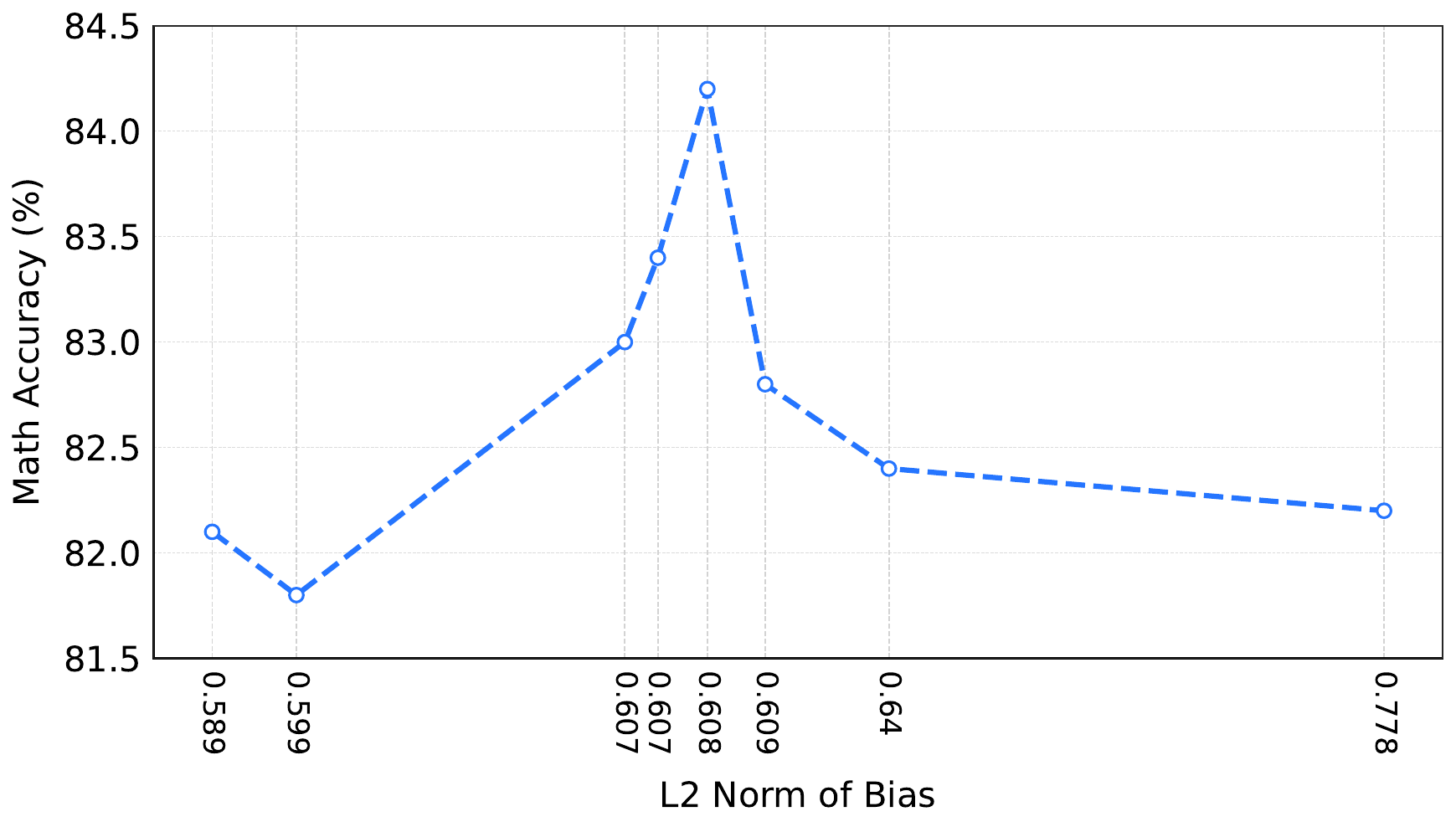} 
\vspace{-10pt}
\caption{The relationship between model performance and bias magnitude.}
\vspace{-10pt}
\label{bias_analyze}
\end{figure}

\subsection{BREP Achieves Optimal Bias Magnitude}

The relationship between $\ell_2$-norm of bias vectors ($\| \mathbf{bias} \|_2$) and model performance is analyzed in this section. As Figure \ref{bias_analyze} illustrates, optimal performance occurs when the $\ell_2$-norm of learned bias vectors converges to a critical threshold of \textbf{0.608}, confirming a strong correlation between bias size and model performance. An excessively small $\ell_2$-norm indicates insufficient influence of the bias term, suggesting underutilization of the ReFT mechanism. Conversely, an excessively large $\ell_2$-norm causes disturbing numerical encoding and error accumulation. Standard ReFT exceeds $\left\| bias_{\text{ReFT}} \right\|_2 = 1.07$, resulting a 6.2\% performance decline in GSM8K. Furthermore, significant variations in optimal vector magnitude are observed in both different model families and differently-sized models within the same family. Notably, BREP employs \textbf{the same target bias within a series of LLM}  (${\textit{target}}_{\text{Llama}}$=$1.0$; ${\textit{target}}_{\text{Qwen}}$=$1.5$), while our PID controller dynamically optimizes $\|bias\|_2$, contributing to the robustness during training. Optimal bias statistical data and train trend are shown in Appendix.

\subsection{BREP Has Stronger Generalization Cabilities}
\vspace{-5pt}
\begin{table}[t]
\centering
\begin{tabular}{c c l l l}
\toprule
\textbf{Model} & \textbf{Method} & \textbf{BoolQ} & \textbf{PIQA} & \textbf{GPQA} \\
\midrule
\multirow{4}{*}{\makecell{\textbf{Llama3}\\\textbf{-8B}\\\textbf{-Instruct}}} 
& \textbf{Base} & 19.4 & 42.4 & 27.0 \\
& \textbf{LoRA} & 18.3\,\textcolor{red}{\scriptsize{↓1.1}} & 47.2\,\textcolor{green}{\scriptsize{↑4.8}} & 21.9\,\textcolor{red}{\scriptsize{↓5.1}} \\
& \textbf{RED} & 15.0\,\textcolor{red}{\scriptsize{↓4.4}} & 63.9\,\textcolor{green}{\scriptsize{↑21.5}} & 24.6\,\textcolor{red}{\scriptsize{↓2.4}} \\
\cmidrule{2-5}
& \textbf{BREP{\scriptsize{(ours)}}} & \textbf{20.7}\,\textcolor{green}{\scriptsize{↑1.3}} & \textbf{60.7}\,\textcolor{green}{\scriptsize{↑18.3}} & \textbf{28.2}\,\textcolor{green}{\scriptsize{↑1.2}} \\
\cmidrule(lr){1-5}
\multirow{4}{*}
{\makecell{\textbf{Qwen3}\\\textbf{-8B}}} 
& \textbf{Base} & 68.9 & 40.1 & 48.7 \\
& \textbf{LoRA} & 69.3\,\textcolor{green}{\scriptsize{↑0.4}} & 40.6\,\textcolor{green}{\scriptsize{↑0.5}} & 48.3\,\textcolor{red}{\scriptsize{↓0.4}} \\
& \textbf{RED} & 70.1\,\textcolor{green}{\scriptsize{↑1.2}} & 23.0\,\textcolor{red}{\scriptsize{↓17.1}} & 49.9\,\textcolor{green}{\scriptsize{↑1.2}} \\
\cmidrule{2-5}
& \textbf{BREP{\scriptsize{(ours)}}} & \textbf{69.1}\,\textcolor{green}{\scriptsize{↑0.2}} & \textbf{42.3}\,\textcolor{green}{\scriptsize{↑2.2}} & \textbf{50.2}\,\textcolor{green}{\scriptsize{↑1.5}} \\
\bottomrule
\end{tabular}
\caption{Comparison of generalization performance.}
\vspace{-5pt}
\label{tab:small_generalization}
\end{table}

\paragraph{Generalization Capability.} BREP’s generalization capability was further verified by evaluating models trained on mathematical datasets on out-of-domain commonsense tasks (BoolQ \cite{clark2019boolq} and PIQA \cite{bisk2020piqa}) and specialized science benchmarks GPQA \cite{rein2024gpqa}. As shown in Table \ref{tab:small_generalization}, BREP achieves the most significant cross-task generalization, delivering the strongest average performance gains across all model architectures. LoRA shows slight improvements ($<1.9\%$) on commonsense tasks but suffers catastrophic degradation on GPQA. RED exhibits high instability on two tasks. Although both BREP and RED enhance model performance through representation manipulation, there are significant differences in their generalization abilities. This phenomenon may stem from smaller vector magnitudes map representations into a broader task subspace, which enhances the model's reasoning and thinking capabilities rather than optimizing task-specific performance.

\begin{figure}[h]
\centering
\vspace{-10pt}
\includegraphics[width=0.95\columnwidth]{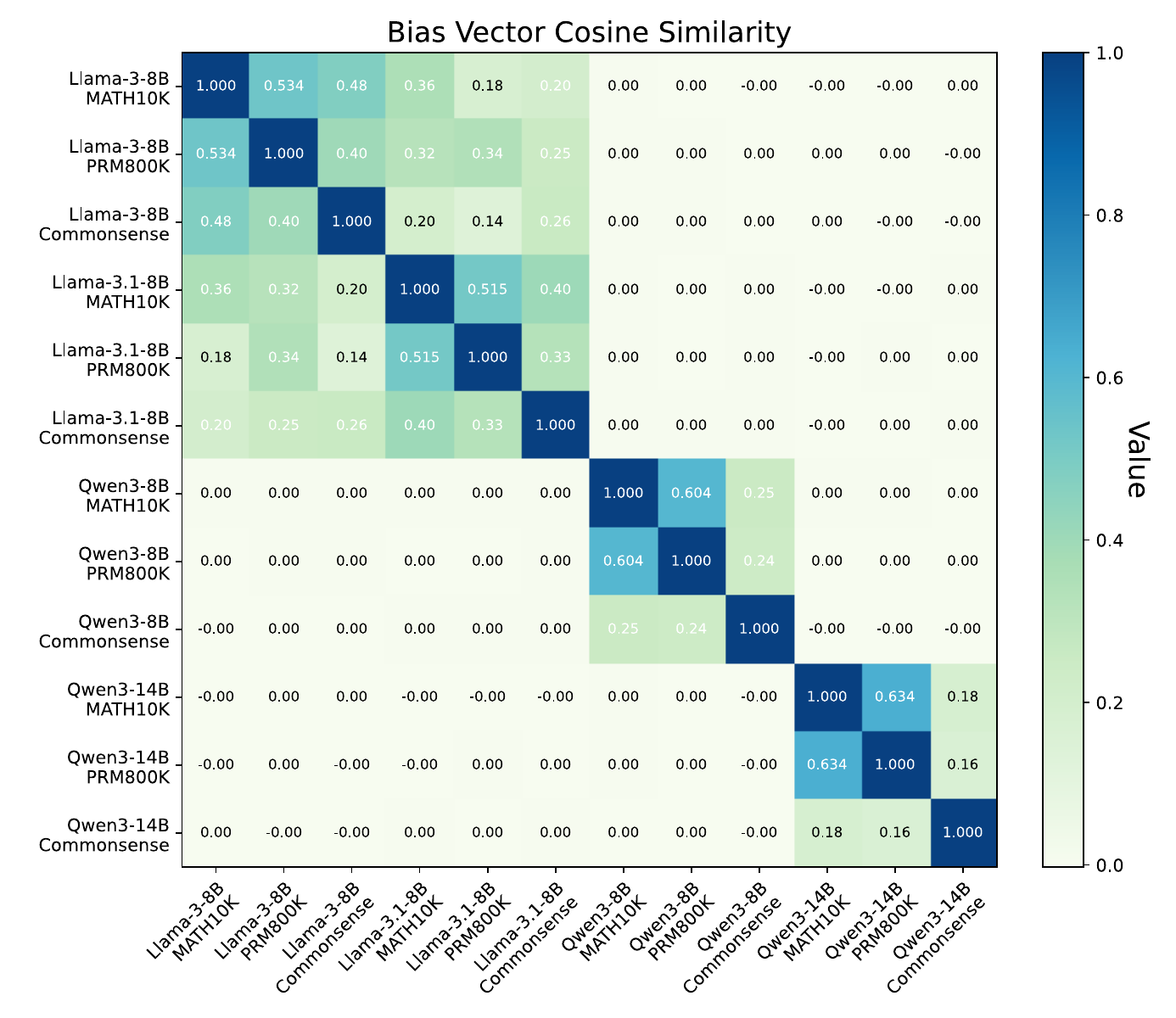} 
\vspace{-10pt}
\caption{Bias cosine similarity between model architecture and tasks.}
\label{samll_bias_analyze}
\vspace{-10pt}
\end{figure}

\paragraph{Architectural Transferability.}  BREP’s transferability across model architecture and tasks is further assessed by measuring bias cosine similarity. As shown in Figure \ref{samll_bias_analyze}, cross-architectural comparisons reveal near-orthogonal relationships, indicating non-transferable representations across model families. Within individual models, vectors exhibit higher similarity for related tasks (the cosine similarity between mathematical reasoning of different difficulties exceeds that between mathematical and commonsense reasoning). These findings confirm that BREP representations are architecture-specific, yet exhibit systematic task-dependent similarities within unified model architecture.

\section{Related Work}

\textbf{Representation Engineering.} Representation Engineering emerged as a paradigm to enhancing the performance, controllability, and efficiency of LLMs by directly manipulating their internal representations \cite{mikolov2013linguistic, liu2023aligning, singh2024mimic, park2023linear, nanda2023emergent}. Unlike traditional weight-based PEFT methods \cite{zaken2021bitfit, liu2023parameter, zhang2023increlora, zhang2024llama}, ReFT treats representations (\textit{e.g.}, hidden states) as fundamental units of analysis and control. Current research focuses on two directions. \textbf{Representation Editing} such as RED \cite{wu2024advancing} modify representations to reduce trainable parameters while maintaining performance on several tasks. LoReFT \cite{wu2024reft} extend this via low-rank subspace interventions, achieving greater parameter efficiency. CRFT \cite{huang2025enhancing} further identifies and optimizes critical representations. \textbf{Cognitive Phenomenon Modeling} like RepE \cite{zouenhancing} enables monitoring and steering high-level behaviors (\textit{e.g.}, truthfulness, emotions) by manipulating vectors. Honesty vectors are used to detect hallucinations \cite{li2023inference}. Overall, representation engineering offers more lightweight and interpretable methods for enhancing the performance of LLM by directly manipulating their internal representations.

\paragraph{Prefix Finetuning.} Recent studies reveal that the initial tokens in LLM responses carry critical signals that influence model behavior \cite{lin2023unlocking, kumar2025detecting}. \citet{qi2025shallow} define shallow preference signals, demonstrating that training on truncated responses achieves competitive or superior reward modeling and DPO performance, suggesting alignment methods often prioritize early tokens over holistic response quality. \citet{ji2025first} extend this insight to reasoning tasks, confirming that early tokens anchor the model’s problem-solving approach, finetuning exclusively on prefixes preserves reasoning efficacy while reducing computational costs by 75-99\%. These works establish that prefix signals govern preference learning and reasoning in LLMs, advocating for more efficient and effective alignment strategies.

\section{Conclusion}

This work addresses the significant performance degradation of ReFT on mathematical tasks. Our analysis reveals misleading reasoning prefixes and disturbing numerical encoding are two primary causes for this problem. Based on this, this paper proposes Bias-Restrained Prefix Representation Finetuning, a novel framework that achieves superior performance to both ReFT and weight-based PEFT methods across multiple mathematical benchmarks while maintaining strong generalization capabilities on several tasks. This work highlights the potential for implementing lighter, more parameter-efficient, and more interpretable ReFT approaches in mathematical reasoning tasks.

\section{Acknowledgments}
This work was supported by Beijing Natural Science Foundation (L243006) and the National Natural Science Foundation of China (No.62406321). This work was also supported by the CCF-NetEase ThunderFire Innovation Research Funding (NO.202507).

\bibliography{main}

@article{wang2025survey,
  title={A survey of recent advances in commonsense knowledge acquisition: Methods and resources},
  author={Wang, Chenhao and Li, Jiachun and Chen, Yubo and Liu, Kang and Zhao, Jun},
  journal={Machine Intelligence Research},
  volume={22},
  number={2},
  pages={201--218},
  year={2025},
  publisher={Springer}
}

@article{xu2025prompting,
  title={Prompting large language models for automatic question tagging},
  author={Xu, Nuojia and Xue, Dizhan and Qian, Shengsheng and Fang, Quan and Hu, Jun},
  journal={Machine Intelligence Research},
  pages={1--12},
  year={2025},
  publisher={Springer}
}

@article{cao2024life,
  title={The life cycle of knowledge in big language models: A survey},
  author={Cao, Boxi and Lin, Hongyu and Han, Xianpei and Sun, Le},
  journal={Machine Intelligence Research},
  volume={21},
  number={2},
  pages={217--238},
  year={2024},
  publisher={Springer}
}

@article{zhang2025crmr,
  title={CRMR: A Collaborative Multistep Reasoning Framework for Solving Mathematical Problems},
  author={Zhang, Yudi and Tang, Xuesong and Hao, Kuangrong},
  journal={Machine Intelligence Research},
  pages={1--14},
  year={2025},
  publisher={Springer}
}

@inproceedings{kumar2025detecting,
  title={Detecting Prefix Bias in LLM-based Reward Models},
  author={Kumar, Ashwin and He, Yuzi and Markosyan, Aram H and Chern, Bobbie and Arrieta-Ibarra, Imanol},
  booktitle={Proceedings of the 2025 ACM Conference on Fairness, Accountability, and Transparency},
  pages={3196--3206},
  year={2025}
}

@article{huang2025enhancing,
  title={Enhancing Chain-of-Thought Reasoning with Critical Representation Fine-tuning},
  author={Huang, Chenxi and Yan, Shaotian and Xie, Liang and Lin, Binbin and Fan, Sinan and Xin, Yue and Cai, Deng and Shen, Chen and Ye, Jieping},
  journal={arXiv preprint arXiv:2507.10085},
  year={2025}
}

@inproceedings{zhang2024llama,
  title={LLaMA-adapter: Efficient fine-tuning of large language models with zero-initialized attention},
  author={Zhang, Renrui and Han, Jiaming and Liu, Chris and Zhou, Aojun and Lu, Pan and Qiao, Yu and Li, Hongsheng and Gao, Peng},
  booktitle={The Twelfth International Conference on Learning Representations},
  year={2024}
}

@article{zhang2023increlora,
  title={Increlora: Incremental parameter allocation method for parameter-efficient fine-tuning},
  author={Zhang, Feiyu and Li, Liangzhi and Chen, Junhao and Jiang, Zhouqiang and Wang, Bowen and Qian, Yiming},
  journal={arXiv preprint arXiv:2308.12043},
  year={2023}
}

@article{liu2023parameter,
  title={Parameter-efficient orthogonal finetuning via butterfly factorization},
  author={Liu, Weiyang and Qiu, Zeju and Feng, Yao and Xiu, Yuliang and Xue, Yuxuan and Yu, Longhui and Feng, Haiwen and Liu, Zhen and Heo, Juyeon and Peng, Songyou and others},
  journal={arXiv preprint arXiv:2311.06243},
  year={2023}
}

@inproceedings{mikolov2013linguistic,
  title={Linguistic regularities in continuous space word representations},
  author={Mikolov, Tom{\'a}{\v{s}} and Yih, Wen-tau and Zweig, Geoffrey},
  booktitle={Proceedings of the 2013 conference of the north american chapter of the association for computational linguistics: Human language technologies},
  pages={746--751},
  year={2013}
}

@article{nanda2023emergent,
  title={Emergent linear representations in world models of self-supervised sequence models},
  author={Nanda, Neel and Lee, Andrew and Wattenberg, Martin},
  journal={arXiv preprint arXiv:2309.00941},
  year={2023}
}

@article{park2023linear,
  title={The linear representation hypothesis and the geometry of large language models},
  author={Park, Kiho and Choe, Yo Joong and Veitch, Victor},
  journal={arXiv preprint arXiv:2311.03658},
  year={2023}
}

@article{singh2024mimic,
  title={Mimic: Minimally modified counterfactuals in the representation space},
  author={Singh, Shashwat and Ravfogel, Shauli and Herzig, Jonathan and Aharoni, Roee and Cotterell, Ryan and Kumaraguru, Ponnurangam},
  journal={Computer Research Repository},
  year={2024}
}

@inproceedings{bisk2020piqa,
  title={Piqa: Reasoning about physical commonsense in natural language},
  author={Bisk, Yonatan and Zellers, Rowan and Gao, Jianfeng and Choi, Yejin and others},
  booktitle={AAAI conference on artificial intelligence (AAAI)},
  volume={34},
  number={05},
  pages={7432--7439},
  year={2020}
}

@article{clark2019boolq,
  title={Boolq: Exploring the surprising difficulty of natural yes/no questions},
  author={Clark, Christopher and Lee, Kenton and Chang, Ming-Wei and Kwiatkowski, Tom and Collins, Michael and Toutanova, Kristina},
  journal={arXiv preprint arXiv:1905.10044},
  year={2019}
}

@inproceedings{rein2024gpqa,
  title={Gpqa: A graduate-level google-proof q\&a benchmark},
  author={Rein, David and Hou, Betty Li and Stickland, Asa Cooper and Petty, Jackson and Pang, Richard Yuanzhe and Dirani, Julien and Michael, Julian and Bowman, Samuel R},
  booktitle={Language Modeling},
  year={2024}
}

@article{belinkov2022probing,
  title={Probing classifiers: Promises, shortcomings, and advances},
  author={Belinkov, Yonatan},
  journal={Computational Linguistics},
  year={2022},
}

@article{liu2023aligning,
  title={Aligning large language models with human preferences through representation engineering},
  author={Liu, Wenhao and Wang, Xiaohua and Wu, Muling and Li, Tianlong and Lv, Changze and Ling, Zixuan and Zhu, Jianhao and Zhang, Cenyuan and Zheng, Xiaoqing and Huang, Xuanjing},
  journal={arXiv preprint arXiv:2312.15997},
  year={2023}
}

@article{aghajanyan2020intrinsic,
  title={Intrinsic dimensionality explains the effectiveness of language model fine-tuning},
  author={Aghajanyan, Armen and Zettlemoyer, Luke and Gupta, Sonal},
  journal={arXiv preprint arXiv:2012.13255},
  year={2020}
}

@article{ding2022delta,
  title={Delta tuning: A comprehensive study of parameter efficient methods for pre-trained language models},
  author={Ding, Ning and Qin, Yujia and Yang, Guang and Wei, Fuchao and Yang, Zonghan and Su, Yusheng and Hu, Shengding and Chen, Yulin and Chan, Chi-Min and Chen, Weize and others},
  journal={arXiv preprint arXiv:2203.06904},
  year={2022}
}

@article{ding2023sparse,
  title={Sparse low-rank adaptation of pre-trained language models},
  author={Ding, Ning and Lv, Xingtai and Wang, Qiaosen and Chen, Yulin and Zhou, Bowen and Liu, Zhiyuan and Sun, Maosong},
  journal={arXiv preprint arXiv:2311.11696},
  year={2023}
}

@article{he2021towards,
  title={Towards a unified view of parameter-efficient transfer learning},
  author={He, Junxian and Zhou, Chunting and Ma, Xuezhe and Berg-Kirkpatrick, Taylor and Neubig, Graham},
  journal={arXiv preprint arXiv:2110.04366},
  year={2021}
}

@article{asai2022attempt,
  title={Attempt: Parameter-efficient multi-task tuning via attentional mixtures of soft prompts},
  author={Asai, Akari and Salehi, Mohammadreza and Peters, Matthew E and Hajishirzi, Hannaneh},
  journal={arXiv preprint arXiv:2205.11961},
  year={2022}
}

@article{tohma2025smartcontrol,
  title={SmartControl: Interactive PID controller design powered by LLM agents and control system expertise},
  author={Tohma, Kadir and Okur, Halil {\.I}brahim and G{\"u}rsoy-Demir, Handan and Ayd{\i}n, Merve Nilay and Yero{\u{g}}lu, Celaleddin},
  journal={SoftwareX},
  year={2025}
}

@book{aastrom2006advanced,
  title={Advanced PID control},
  author={{\AA}str{\"o}m, Karl Johan and H{\"a}gglund, Tore},
  year={2006},
  publisher={ISA-The Instrumentation, Systems and Automation Society}
}

@article{cobbe2021training,
  title={Training verifiers to solve math word problems},
  author={Cobbe, Karl and Kosaraju, Vineet and Bavarian, Mohammad and Chen, Mark and Jun, Heewoo and Kaiser, Lukasz and Plappert, Matthias and Tworek, Jerry and Hilton, Jacob and Nakano, Reiichiro and others},
  journal={arXiv preprint arXiv:2110.14168},
  year={2021}
}

@article{wei2022chain,
  title={Chain-of-thought prompting elicits reasoning in large language models},
  author={Wei, Jason and Wang, Xuezhi and Schuurmans, Dale and Bosma, Maarten and Xia, Fei and Chi, Ed and Le, Quoc V and Zhou, Denny and others},
  journal={Advances in Neural Information Processing Systems (NeurIPS)},
  volume={35},
  pages={24824--24837},
  year={2022}
}

@article{li2021prefix,
  title={Prefix-tuning: Optimizing continuous prompts for generation},
  author={Li, Xiang Lisa and Liang, Percy},
  journal={arXiv preprint arXiv:2101.00190},
  year={2021}
}

@inproceedings{houlsby2019parameter,
  title={Parameter-efficient transfer learning for NLP},
  author={Houlsby, Neil and Giurgiu, Andrei and Jastrzebski, Stanislaw and Morrone, Bruna and De Laroussilhe, Quentin and Gesmundo, Andrea and Attariyan, Mona and Gelly, Sylvain},
  booktitle={International Conference on Machine Learning (ICML)},
  pages={2790--2799},
  year={2019},
  organization={PMLR}
}

@article{achiam2023gpt,
  title={Gpt-4 technical report},
  author={Achiam, Josh and Adler, Steven and Agarwal, Sandhini and Ahmad, Lama and Akkaya, Ilge and Aleman, Florencia Leoni and Almeida, Diogo and Altenschmidt, Janko and Altman, Sam and Anadkat, Shyamal and others},
  journal={arXiv preprint arXiv:2303.08774},
  year={2023}
}

@article{anil2023palm,
  title={Palm 2 technical report},
  author={Anil, Rohan and Dai, Andrew M and Firat, Orhan and Johnson, Melvin and Lepikhin, Dmitry and Passos, Alexandre and Shakeri, Siamak and Taropa, Emanuel and Bailey, Paige and Chen, Zhifeng and others},
  journal={arXiv preprint arXiv:2305.10403},
  year={2023}
}

@article{yang2025qwen3,
  title={Qwen3 technical report},
  author={Yang, An and Li, Anfeng and Yang, Baosong and Zhang, Beichen and Hui, Binyuan and Zheng, Bo and Yu, Bowen and Gao, Chang and Huang, Chengen and Lv, Chenxu and others},
  journal={arXiv preprint arXiv:2505.09388},
  year={2025}
}

@article{bai2023qwen,
  title={Qwen technical report},
  author={Bai, Jinze and Bai, Shuai and Chu, Yunfei and Cui, Zeyu and Dang, Kai and Deng, Xiaodong and Fan, Yang and Ge, Wenbin and Han, Yu and Huang, Fei and others},
  journal={arXiv preprint arXiv:2309.16609},
  year={2023}
}

@article{grattafiori2024llama,
  title={The llama 3 herd of models},
  author={Grattafiori, Aaron and Dubey, Abhimanyu and Jauhri, Abhinav and Pandey, Abhinav and Kadian, Abhishek and Al-Dahle, Ahmad and Letman, Aiesha and Mathur, Akhil and Schelten, Alan and Vaughan, Alex and others},
  journal={arXiv preprint arXiv:2407.21783},
  year={2024}
}

@article{dai2015semi,
  title={Semi-supervised sequence learning},
  author={Dai, Andrew M and Le, Quoc V},
  journal={Advances in Neural Information Processing Systems (NeurIPS)},
  volume={28},
  year={2015}
}

@article{hendrycks2021measuring,
  title={Measuring mathematical problem solving with the math dataset},
  author={Hendrycks, Dan and Burns, Collin and Kadavath, Saurav and Arora, Akul and Basart, Steven and Tang, Eric and Song, Dawn and Steinhardt, Jacob},
  journal={arXiv preprint arXiv:2103.03874},
  year={2021}
}

@article{amini2019mathqa,
  title={Mathqa: Towards interpretable math word problem solving with operation-based formalisms},
  author={Amini, Aida and Gabriel, Saadia and Lin, Peter and Koncel-Kedziorski, Rik and Choi, Yejin and Hajishirzi, Hannaneh},
  journal={arXiv preprint arXiv:1905.13319},
  year={2019}
}

@article{patel2021nlp,
  title={Are NLP models really able to solve simple math word problems?},
  author={Patel, Arkil and Bhattamishra, Satwik and Goyal, Navin},
  journal={arXiv preprint arXiv:2103.07191},
  year={2021}
}

@article{hu2023llm,
  title={Llm-adapters: An adapter family for parameter-efficient fine-tuning of large language models},
  author={Hu, Zhiqiang and Wang, Lei and Lan, Yihuai and Xu, Wanyu and Lim, Ee-Peng and Bing, Lidong and Xu, Xing and Poria, Soujanya and Lee, Roy Ka-Wei},
  journal={arXiv preprint arXiv:2304.01933},
  year={2023}
}

@inproceedings{zhu2025language,
  title={Language models encode the value of numbers linearly},
  author={Zhu, Fangwei and Dai, Damai and Sui, Zhifang},
  booktitle={International Conference on Computational Linguistics},
  pages={693--709},
  year={2025}
}

@inproceedings{lightman2023let,
  title={Let's verify step by step},
  author={Lightman, Hunter and Kosaraju, Vineet and Burda, Yuri and Edwards, Harrison and Baker, Bowen and Lee, Teddy and Leike, Jan and Schulman, John and Sutskever, Ilya and Cobbe, Karl},
  booktitle={International Conference on Learning Representations (ICLR)},
  year={2023}
}

@article{qi2025shallow,
  title={Shallow Preference Signals: Large Language Model Aligns Even Better with Truncated Data?},
  author={Qi, Xuan and Qiu, Jiahao and Juan, Xinzhe and Wu, Yue and Wang, Mengdi},
  journal={arXiv preprint arXiv:2505.17122},
  year={2025}
}

@article{ji2025first,
  title={The first few tokens are all you need: An efficient and effective unsupervised prefix fine-tuning method for reasoning models},
  author={Ji, Ke and Xu, Jiahao and Liang, Tian and Liu, Qiuzhi and He, Zhiwei and Chen, Xingyu and Liu, Xiaoyuan and Wang, Zhijie and Chen, Junying and Wang, Benyou and others},
  journal={arXiv preprint arXiv:2503.02875},
  year={2025}
}

@article{lin2023unlocking,
  title={The unlocking spell on base llms: Rethinking alignment via in-context learning},
  author={Lin, Bill Yuchen and Ravichander, Abhilasha and Lu, Ximing and Dziri, Nouha and Sclar, Melanie and Chandu, Khyathi and Bhagavatula, Chandra and Choi, Yejin},
  journal={arXiv preprint arXiv:2312.01552},
  year={2023}
}

@article{zouenhancing,
  title={Enhancing Neural Network Transparency through Representation Analysis},
  author={Zou, Andy and Phan, Long and Chen, Sarah Li and Campbell, James and Guo, Phillip Huang and Ren, Richard and Pan, Alexander and Yin, Xuwang and Mazeika, Mantas and Dombrowski, Annah and others},
    journal={arXiv preprint},
  year={2024}
}

@article{hojer2025improving,
  title={Improving reasoning performance in large language models via representation engineering},
  author={H{\o}jer, Bertram and Jarvis, Oliver and Heinrich, Stefan},
  journal={arXiv preprint arXiv:2504.19483},
  year={2025}
}

@article{tang2025unlocking,
  title={Unlocking General Long Chain-of-Thought Reasoning Capabilities of Large Language Models via Representation Engineering},
  author={Tang, Xinyu and Wang, Xiaolei and Lv, Zhihao and Min, Yingqian and Zhao, Wayne Xin and Hu, Binbin and Liu, Ziqi and Zhang, Zhiqiang},
  journal={arXiv preprint arXiv:2503.11314},
  year={2025}
}

@article{wu2024advancing,
  title={Advancing parameter efficiency in fine-tuning via representation editing},
  author={Wu, Muling and Liu, Wenhao and Wang, Xiaohua and Li, Tianlong and Lv, Changze and Ling, Zixuan and Zhu, Jianhao and Zhang, Cenyuan and Zheng, Xiaoqing and Huang, Xuanjing},
  journal={arXiv preprint arXiv:2402.15179},
  year={2024}
}

@article{wu2024reft,
  title={Reft: Representation finetuning for language models},
  author={Wu, Zhengxuan and Arora, Aryaman and Wang, Zheng and Geiger, Atticus and Jurafsky, Dan and Manning, Christopher D and Potts, Christopher},
  journal={Advances in Neural Information Processing Systems (NeurIPS)},
  year={2024}
}

@article{li2023inference,
  title={Inference-time intervention: Eliciting truthful answers from a language model},
  author={Li, Kenneth and Patel, Oam and Vi{\'e}gas, Fernanda and Pfister, Hanspeter and Wattenberg, Martin},
  journal={Advances in Neural Information Processing Systems (NeurIPS)},
  year={2023}
}

@article{hu2022lora,
  title={Lora: Low-rank adaptation of large language models.},
  author={Hu, Edward J and Shen, Yelong and Wallis, Phillip and Allen-Zhu, Zeyuan and Li, Yuanzhi and Wang, Shean and Wang, Lu and Chen, Weizhu and others},
  journal={International Conference on Learning Representations (ICLR)},
  year={2022}
}

@article{brown2020language,
  title={Language models are few-shot learners},
  author={Brown, Tom and Mann, Benjamin and Ryder, Nick and Subbiah, Melanie and Kaplan, Jared D and Dhariwal, Prafulla and Neelakantan, Arvind and Shyam, Pranav and Sastry, Girish and Askell, Amanda and others},
  journal={Advances in Neural Information Processing Systems (NeurIPS)},
  volume={33},
  pages={1877--1901},
  year={2020}
}


\definecolor{purple}{RGB}{0,0,0}
\definecolor{blue}{RGB}{0,0,0}
\definecolor{green}{rgb}{0.0, 0.0, 0.0}
\definecolor{red}{rgb}{0.0, 0.0, 0.0}

\lstset{%
	basicstyle={\footnotesize\ttfamily},
	numbers=left,numberstyle=\footnotesize,xleftmargin=2em,
	aboveskip=0pt,belowskip=0pt,%
	showstringspaces=false,tabsize=2,breaklines=true}
\floatstyle{ruled}
\newfloat{listing}{tb}{lst}{}
\floatname{listing}{Listing}
%
\pdfinfo{
/TemplateVersion (2026.1)
}

\setcounter{secnumdepth}{0} 

%


\title{Appendix}
\maketitle

\section{Appendix}

\begin{figure}[htbp]
\centering
\includegraphics[width=1\columnwidth]{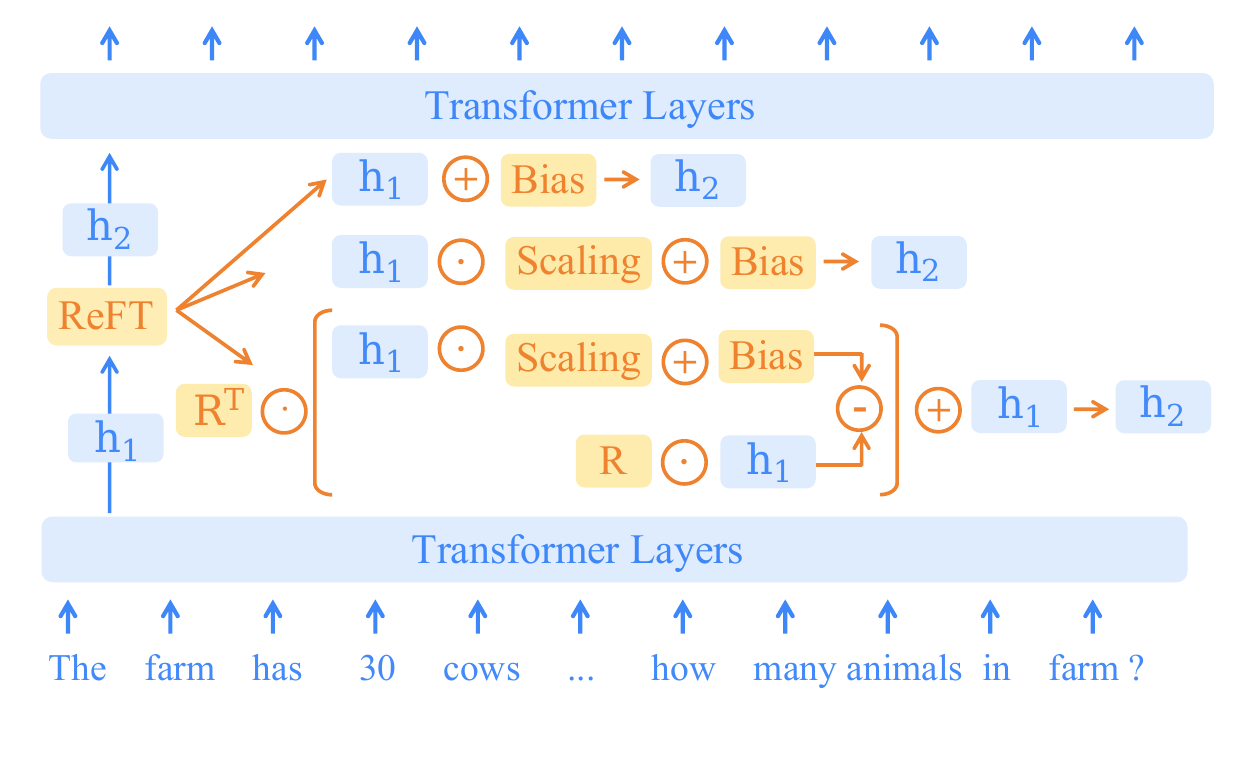} 
\caption{Bias cosine similarity between model architecture and tasks.}
\label{ReFT}
\vspace{-10pt}
\end{figure}

Representation Fine-Tuning (ReFT) is a recently proposed parameter-efficient fine-tuning paradigm that enhances model performance by modifying internal representations while keeping the model weights fixed. Figure \ref{ReFT} shows three common methods of ReFT. In contrast to conventional PEFT methods such as LoRA or Adapter, which inject additional trainable parameters into model layers, ReFT applies lightweight learnable transformations—typically scaling and bias operations—directly on the hidden states across layers. These modifications enable efficient adaptation to downstream tasks with significantly fewer trainable parameters. Prior work has demonstrated that ReFT achieves competitive or superior performance in tasks such as commonsense reasoning and instruction following. However, empirical results indicate that ReFT struggles on complex, multi-step reasoning tasks like mathematical problem solving, where error accumulation and representation corruption can severely impair accuracy.

\begin{figure}[htbp]
\centering
\includegraphics[width=1\columnwidth]{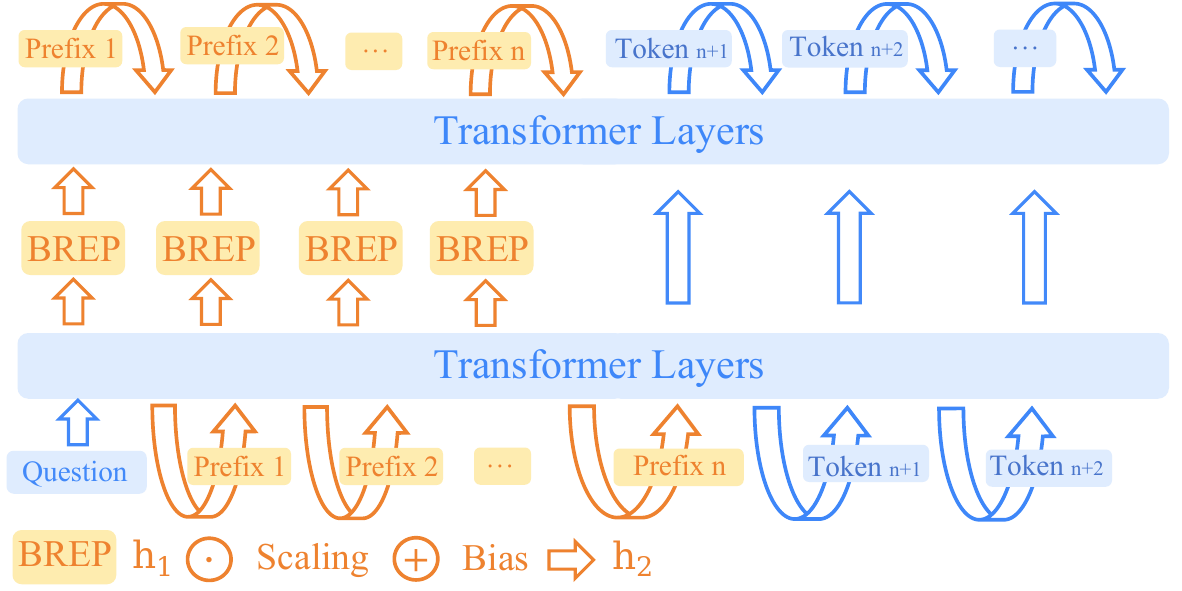} 
\caption{Bias cosine similarity between model architecture and tasks.}
\label{BREP}
\vspace{-10pt}
\end{figure}

To address the limitations of ReFT in mathematical reasoning, we propose Bias-Restrained Prefix Representation Fine-Tuning (BREP), a novel ReFT-based framework designed to enhance both reasoning accuracy and numerical fidelity (shown in Figure \ref{BREP}). BREP introduces two key innovations: (1) Prefix Truncation with Early-Stage Intervention, which focuses training and inference interventions on the initial tokens of reasoning sequences, thereby improving the quality of reasoning prefixes while mitigating downstream error propagation; and (2) Bias Constraint Training, which employs a PID-based controller to dynamically regulate the magnitude of intervention vectors, preventing numerical encoding distortions caused by over- or under-specified biases. Extensive experiments across multiple LLM architectures and mathematical benchmarks demonstrate that BREP consistently outperforms both standard ReFT and weight-based PEFT methods, achieving stronger performance on complex chain-of-thought tasks while preserving generalization to out-of-domain settings.
 
\section{Preliminary Supplement}

\subsection{Prefix Guiding Effect}

\begin{figure*}[h]
\centering
\includegraphics[width=2.1\columnwidth]{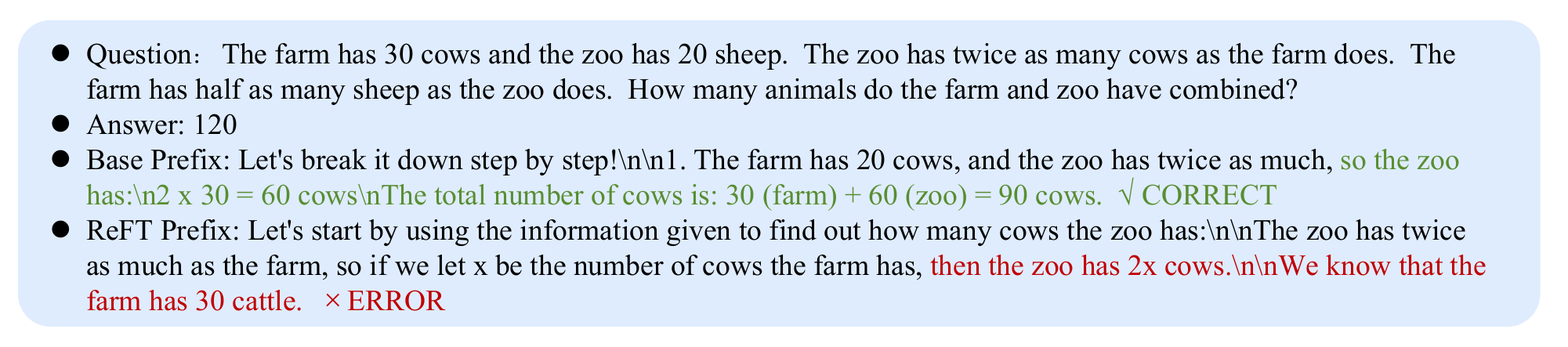} 
\caption{A comparasion of Base prefix and ReFT prefix.}
\label{Prefx}
\end{figure*}

Figure \ref{Prefx} shows an example of how the ReFT and Base prefixes guide the model to answering the questions.

\subsection{Number Probe Train}
A synthetic dataset of 9,000 addition problems was generated, formatted as Table \ref{tab:number_data}.

\begin{table}[htbp]
\centering
\begin{tabular}{l}
\toprule
\textbf{Question}: What is the sum of \{a\} and \{b\}? \\
\textbf{Answer}: \{a+b\} \\
\bottomrule
\end{tabular}
\caption{Number probe train data example}
\label{tab:number_data}
\vspace{-0.2cm}
\smallskip
\end{table}

$a$ and $b$ are randomly generated integers with digit lengths ranging from 2 to 10. The dataset was partitioned into training (80\%), validation (10\%), and test (10\%) sets. Hidden states $h_i \in \mathbb{R}^{d_{\text{model}}}$ were extracted from transformer layers of pretrained LLMs (Llama-3-8B-Instruct) during forward passes of question texts. States corresponding to three critical positions were analyzed: Last token of first number ($a$), Last token of second number ($b$), and Last token of input text ($o$).

A Ridge regression model with L2 regularization ($\lambda = 0.1$) was implemented:
$$
\min_{W,b} \left\| \log_2(X) - H W - b \right\|_2^2 + \lambda \|W\|_2^2
$$
where $H$ denotes hidden state matrix, $X$ denotes target numerical values and $W \in \mathbb{R}^{d_{\text{model}}}, b$ are learnable parameters. Logarithmic transformation $\log_2(x)$ ensured numerical stability during training.

The computation of Pearson coefficients for the probes revealed a significant pattern across different model layers and token positions. As illustrated in Figure \ref{numprob_preason}, the Pearson coefficients at all three critical probed positions—namely, the final token of the first input number (a), the final token of the second input number (b), and the final token of the input text (o)—consistently approach 1. This strong linear correlation provides evidence for the linear encoding characteristics of numerical values within language models' hidden representations.

\begin{table*}[htbp]
\centering
\begin{tabular}{l l l c c c}
\toprule
\textbf{Method} & \textbf{Error Type} & \textbf{Detailed Division} & \textbf{Count} & \textbf{Percentage} & \textbf{Total} \\ 
\midrule
\multirow{7}{*}{\textbf{ReFT}} & \multirow{3}{*}{\textbf{Misleading Error}} & Logical error & 45 & 39.5\% & \multirow{3}{*}{54.4\%} \\ \cmidrule{3-5}
 &  & Mistaken understanding & 14 & 12.3\% &  \\ \cmidrule{3-5}
 &  & Introduced virtual data & 3 & 2.6\% &  \\ \cmidrule{2-6}
 & \multirow{2}{*}{\textbf{Numerical encoding error}} & Confuse or forget key data & 21 & 18.4\% & \multirow{2}{*}{45.6\%} \\ \cmidrule{3-5}
 &  & Calculation error & 31 & 27.2\% &  \\ \cmidrule{2-6}
 & \multicolumn{2}{c}{Total} & 114 & 100.0\% & 100.0\% \\ 
 \toprule 
\multirow{7}{*}{\textbf{BREP}} & \multirow{3}{*}{\textbf{Misleading Error}} & Logical error & 30 & 37.0\% & \multirow{3}{*}{64.2\%} \\ \cmidrule{3-5}
 &  & Mistaken understanding & 15 & 18.5\% &  \\ \cmidrule{3-5}
 &  & Introduced virtual data & 7 & 8.6\% &  \\ \cmidrule{2-6}
 & \multirow{2}{*}{\textbf{Numerical encoding error}} & Confuse or forget key data & 11 & 13.6\% & \multirow{2}{*}{35.8\%} \\ \cmidrule{3-5}
 &  & Calculation error & 18 & 22.2\% &  \\ \cmidrule{2-6}
 & \multicolumn{2}{c}{Total} & 81 & 100.0\% & 100.0\% \\ \bottomrule
\end{tabular}
\caption{Error type statistics of ReFT and BREP on Llama3-8B-Instruct.}
\label{tab:error_analysis_combined}
\end{table*}

\begin{figure*}[bp]
\centering
\includegraphics[width=2.1\columnwidth]{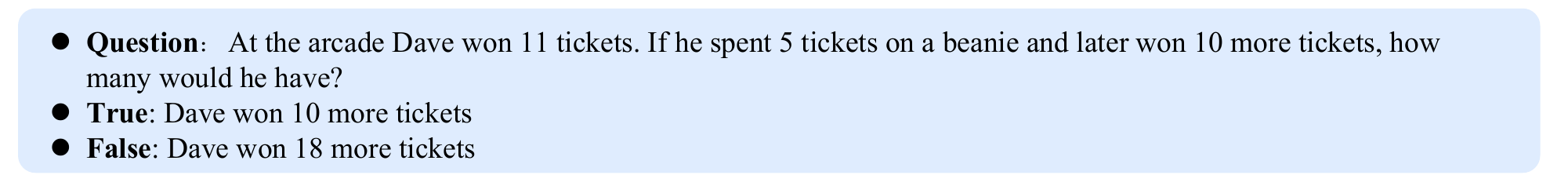} 
\caption{Prob training data example.}
\label{prob_data}
\end{figure*}

\begin{figure*}[h]
\centering
\includegraphics[width=2.1\columnwidth]{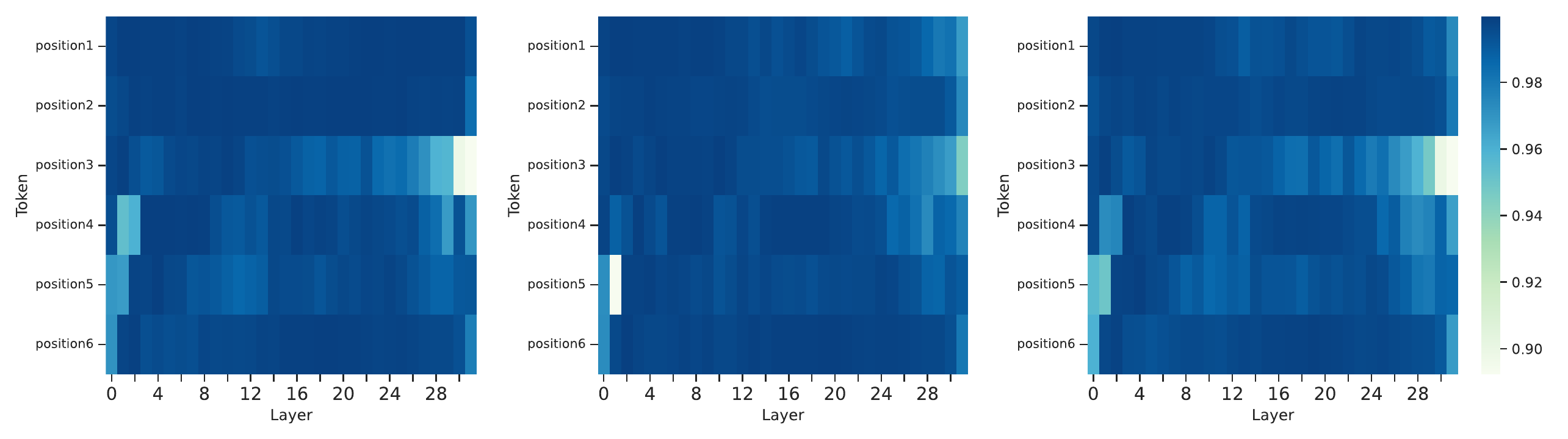} 
\caption{The computation of Pearson coefficients for the probes. The Pearson coefficients at all three critical probed positions: the final token of the first input number (a), the final token of the second input number (b), and the final token of the input text (o).}
\label{numprob_preason}
\end{figure*}

\section{Train Details}

\begin{figure*}[h]
\centering
\includegraphics[width=2.1\columnwidth]{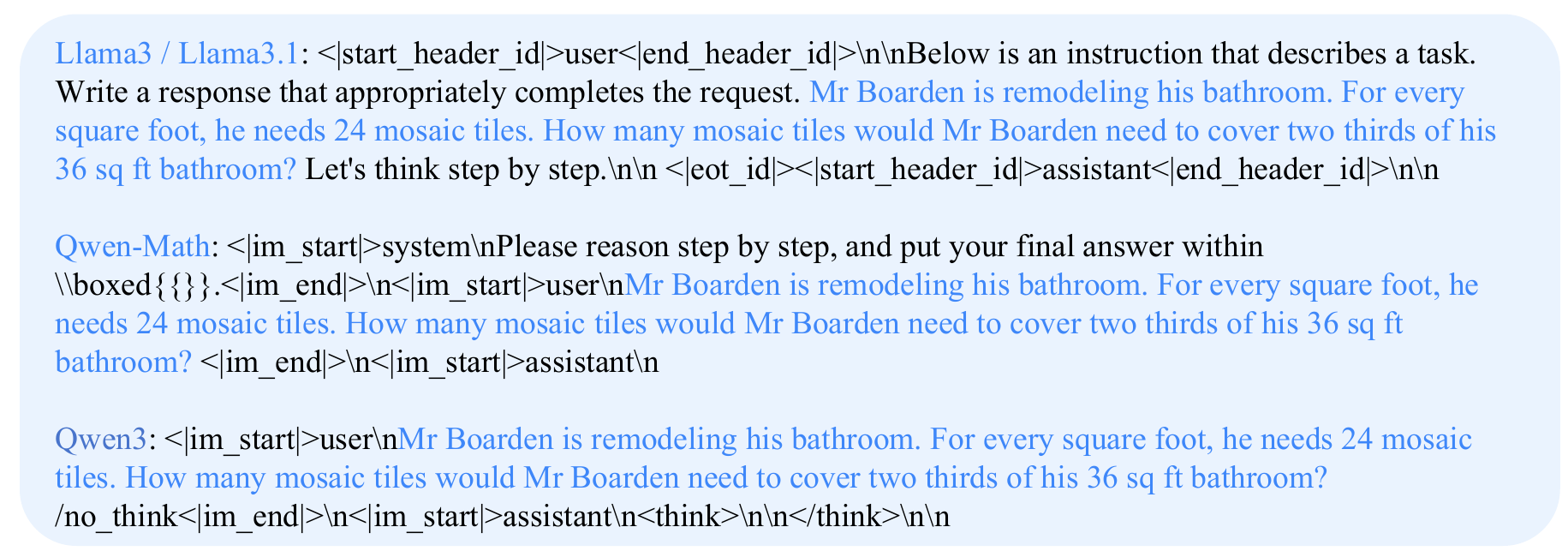} 
\caption{Generation template of Llama3-8B-Instruct, Llama3.1-8B-Instruct, Qwen2.5-Math-7B-Instruct, Qwen3-8B and Qwen3-14B.}
\label{template}
\end{figure*}

Figure \ref{template} shows our question template, we use the prompt method of \textbf{Zero-shot CoT}.

Based on experimental results referenced in Table \ref{tab:prefix_length}, we set the train prefix length to 64 and intervention prefix length to 8. The question template employs Zero-shot CoT prompting as depicted in Figure \ref{template}. For Qwen-series models, we utilize their default generation configurations, while for Llama-series models we employ deterministic greedy decoding and a strict 5-gram repetition constraint to prevent local sequence repetitions.

\begin{figure*}[h]
\centering
\includegraphics[width=2.1\columnwidth]{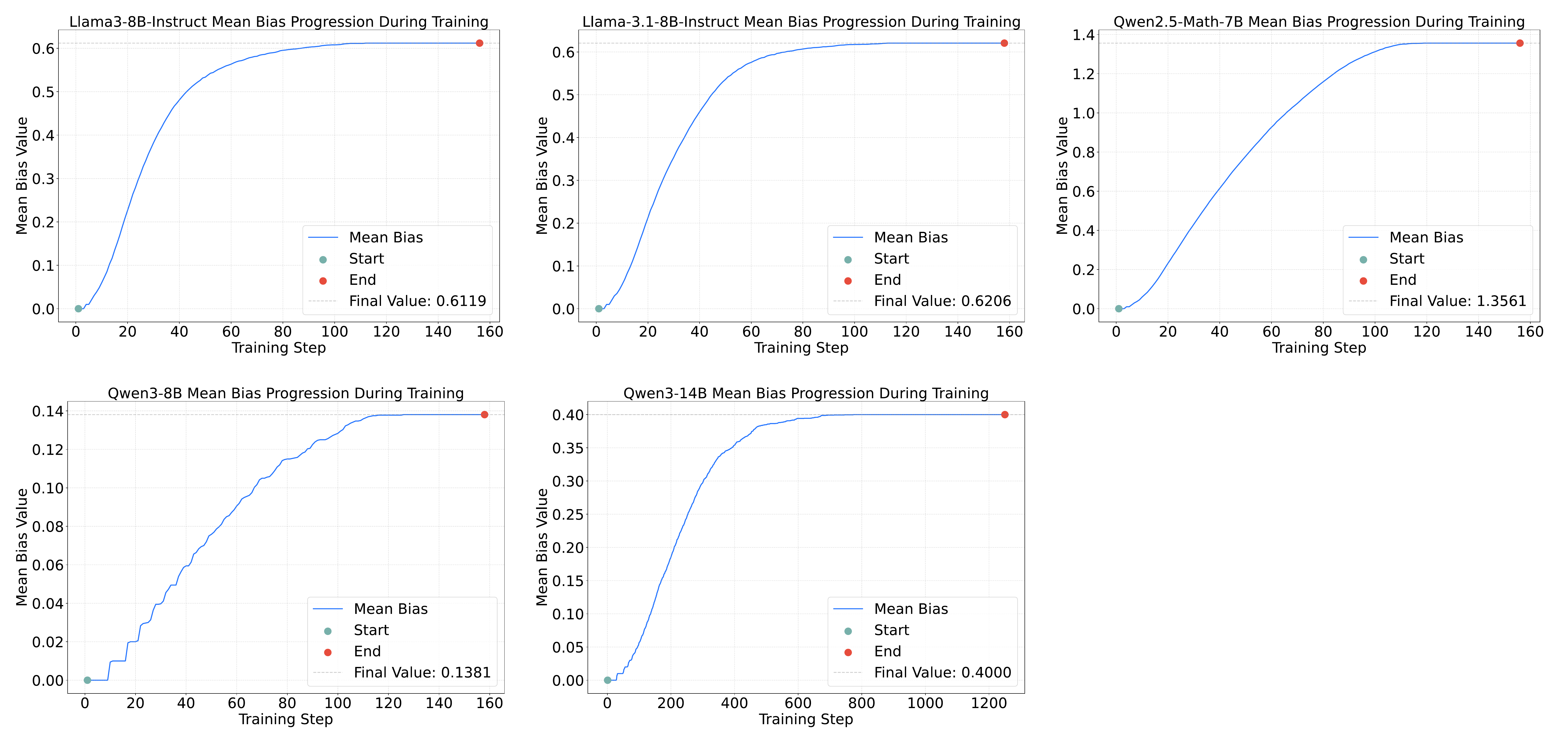} 
\caption{The changing trend of the bias $\ell_2$-norm.}
\label{bias_trend}
\end{figure*}

Figure \ref{bias_trend} shows the bias trend during training, PID can learn quickly in the early stage and slow down the learning speed and achieve the ideal bias length ultimately. Hyperparameter setting is shown in Table \ref{Hyperparameter}.

\begin{table}[ht]
\centering
\begin{tabular}{c c c c}
\toprule
\textbf{Train Prefix} & \textbf{Acc} & \textbf{Intervene Prefix} &\textbf{Acc} \\ 
\midrule
\textbf{8} & 82.4 & \textbf{4} & 82.3\\
\textbf{16} & 81.8 & \textbf{8} & \textbf{82.8}\\ 
\textbf{32} & 82.3 & \textbf{16} & 81.0\\
\textbf{64} & \textbf{82.8} & \textbf{32} & 82.6 \\
\textbf{128} & 78.4 & \textbf{64} & 81.3 \\ 
\textbf{256} & 81.4 & \textbf{128} & 80.7 \\
\textbf{512} & 81.0 & \textbf{256} & 80.7 \\
\bottomrule
\end{tabular}
\caption{The influence of training prefix and intervention prefix on model performance, where a fixed intervention prefix (Intervene Prefix=8) is used when testing training prefix and a fixed training prefix (Train Prefix=64) is used when testing intervention prefix}
\label{tab:prefix_length}
\end{table}

\section{Comparasion Method}
\begin{itemize}
\item \textbf{Low-Rank Adaption (LoRA)} employs a low-rank decomposition on the matrix $\Delta W$, thereby modeling weight updates as the product of two low-rank matrices. These two learnable matrices are aligned in parallel with the corresponding matrices in pre-trained models. They process inputs in parallel and combine their results to generate the final outputs in each transformer block.

\item \textbf{RED} incorporates a learnable scaling vector $l_{scaling}$ and apply it to perform the Hadamard product with a hidden representation $h_1$ by scaling the feature of each dimension within $h_1$ via element-wise multiplication. Additionally, introducing another learnable bias vector $l_{bias}$ that is subsequently added to the scaled vector.

\item \textbf{LoReFT} Low-rank Linear Subspace ReFT is a parameter-efficient finetuning method that operates on a frozen base model by learning task-specific interventions on its hidden representations. Instead of updating the model's weights, LoReFT modifies a representation vector \(\mathbf{h}\) by adding a learned, task-specific edit vector. This edit is constrained to an \(r\)-dimensional linear subspace, which is defined by a low-rank projection matrix \(\mathbf{R} \in \mathbb{R}^{r \times d}\). The intervention is formulated as \(\Phi_{\text{LoReFT}}(\mathbf{h}) = \mathbf{h} + \mathbf{R}^T(\mathbf{W}\mathbf{h} + \mathbf{b} - \mathbf{R}\mathbf{h})\), where the parameters \(\theta = \{\mathbf{R}, \mathbf{W}, \mathbf{b}\}\) are learned to steer the model's behavior towards a downstream task. This approach allows for highly parameter-efficient adaptation, as it directly manipulates semantic information encoded in representations rather than modifying the core model architecture.
\end{itemize}

\begin{table*}[ht]
\centering
\begin{tabular}{c c c c c c}
\toprule
\textbf{Model} & \textbf{Train Prefix} & \textbf{Intervene Prefix} & \textbf{Learning Rate} & \textbf{Target Bias} & \textbf{Learned Bias} \\
\midrule
Llama3-8B-Instruct & 64 & 8 & $2 \times 10^{-4}$ & 1 & 0.611 \\
Llama-3.1-8B-Instruct & 65 & 9 & $2 \times 10^{-4}$ & 1 & 0.621 \\
Qwen2.5-Math-7B & 66 & 10 & $2 \times 10^{-4}$ & 1.5 & 1.355 \\
Qwen3-8B & 67 & 11 & $2 \times 10^{-5}$ & 1.5 & 0.137 \\
Qwen3-14B & 68 & 12 & $2 \times 10^{-5}$ & 1.5 & 0.377 \\
\bottomrule
\end{tabular}
\caption{Hyperparameter config of Llama3-8B-Instruct, Llama-3.1-8B-Instruct, Qwen2.5-Math-7B, Qwen3-8B and Qwen3-14B. We use greedy decoding without sampling for Llama and default setting for Qwen.} 
\label{Hyperparameter}
\end{table*}

\begin{figure*}[h]
\centering
\includegraphics[width=1.5\columnwidth]{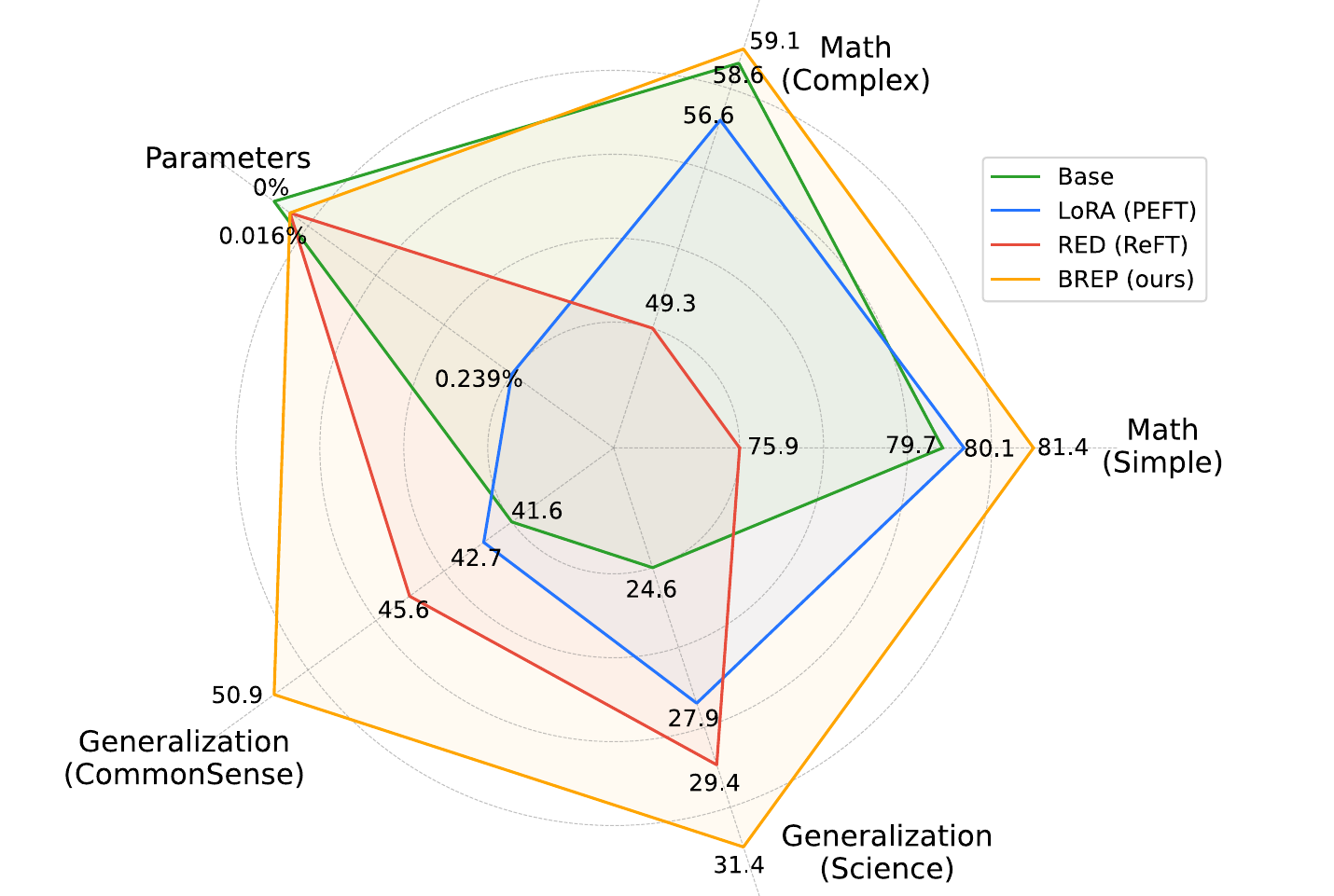} 
\caption{A performance comparison of Base, RED, and BREP.}
\label{rader}
\end{figure*}

\section{Datasets}

The training and test sets we used in this work are as follows:

\begin{itemize}
\item \textbf{MATH10K}: A combined training dataset of four tasks mentioned above: GSM8K, MAWPS, MAWPS-single. Selected tasks are excluded for testing since the original paper accidentally leaks testing examples from these tasks into the training set, affecting AddSub, MultiArith and SingleEq.
\item \textbf{PRM800K}: A complete dataset of 800,000 step-level human feedback labels.
\item \textbf{GSM8K}: A dataset consists of grade-school math word problems that require multi-step reasoning.
\item \textbf{SVAMP}: A dataset enhances the original Math World Problem (MWP) challenge by requiring robust reasoning ability that is invariant to structural alternations of the posing problem.
\item \textbf{MATHQA}: A large-scale dataset of math word problems and an interpretable neural math problem solver that learns to map problems to operation programs.
\item \textbf{MATH500}: A dataset comprises 500 high-school-level mathematical problems.
\item \textbf{AMC23}: a challenging mathematical reasoning benchmark used to evaluate AI models' problem-solving capabilities on complex, competition-level algebra and geometry problems.
\item \textbf{Commonsense170K}:  A combined dataset of eight commonsense reasoning tasks.
\item \textbf{BoolQ}: A question-answering dataset for yes or no naturally occurring questions. We remove the provided passage in the dataset following previous works to ensure a fair comparison.
\item \textbf{PIQA}: A dataset tests physical commonsense reasoning and requires the model to choose one of the provided actions to take based on a hypothesised scenario.
\item \textbf{GPQA}: A challenging dataset of 448 multiple-choice questions written by domain experts in biology, physics, and chemistry.
\item \textbf{Ultrafeedback}:  A dataset for future RLHF and feedback learning research. 
\end{itemize}

\section{Probe Details}

A standard methodology for identifying a model's internal representations is probes, which is training a classifier on the activations of network. We constructed a dataset of 3,000 instances to probe the faithfulness of the model, labeling instances consistent with questions' numerical values as correct and inconsistent instances as incorrect (an example are provided in Figure \ref{prob_data}). Probes were trained to assess accuracy at every layer of the model. Validation confirmed probe reliability, with almost all layers demonstrating probe accuracy exceeding 80\% (shown in Table \ref{tab:prob_acc}).




\begin{table*}[h]
\centering
\begin{tabular}{@{}l *{16}{c} @{}}
\toprule
\multicolumn{16}{c}{\textbf{Layer}} \\
\cmidrule(lr){2-17}
\textbf{Model} & 0 & 1 & 2 & 3 & 4 & 5 & 6 & 7 & 8 & 9 & 10 & 11 & 12 & 13 & 14 & 15 \\
\midrule
\textbf{Base}   & 0.64 & 0.67 & 0.67 & 0.74 & 0.78 & 0.80 & 0.81 & 0.85 & 0.84 & 0.84 & 0.85 & 0.83 & 0.85 & 0.86 & 0.85 & 0.85 \\
\textbf{BREP} & 0.64 & 0.68 & 0.68 & 0.75 & 0.78 & 0.78 & 0.81 & 0.82 & 0.83 & 0.83 & 0.84 & 0.83 & 0.85 & 0.85 & 0.86 & 0.85 \\
\textbf{ReFT}   & 0.64 & 0.70 & 0.71 & 0.73 & 0.78 & 0.79 & 0.83 & 0.85 & 0.84 & 0.85 & 0.84 & 0.84 & 0.86 & 0.85 & 0.85 & 0.86 \\
\toprule
\multicolumn{16}{c}{\textbf{Layer}} \\
\cmidrule(lr){2-17}
\textbf{Model} & 16 & 17 & 18 & 19 & 20 & 21 & 22 & 23 & 24 & 25 & 26 & 27 & 28 & 29 & 30 & 31 \\
\midrule
\textbf{Base}   & 0.85 & 0.86 & 0.85 & 0.86 & 0.85 & 0.86 & 0.86 & 0.86 & 0.86 & 0.87 & 0.86 & 0.86 & 0.84 & 0.84 & 0.84 & 0.85 \\
\textbf{BREP} & 0.85 & 0.86 & 0.86 & 0.85 & 0.85 & 0.83 & 0.83 & 0.83 & 0.83 & 0.84 & 0.84 & 0.84 & 0.82 & 0.82 & 0.81 & 0.80 \\
\textbf{ReFT}   & 0.84 & 0.85 & 0.86 & 0.86 & 0.86 & 0.86 & 0.85 & 0.84 & 0.85 & 0.84 & 0.84 & 0.83 & 0.84 & 0.83 & 0.84 & 0.83 \\
\bottomrule
\end{tabular}
\caption{The accuracy of the probes at each layer of the Base, BREP and ReFT Model.}
\label{tab:prob_acc}
\end{table*}

\begin{table*}[tbp]
\centering
\begin{tabular}{@{}c *{16}{c} @{}}
\toprule
& \multicolumn{16}{c}{\textbf{Layer}} \\
\cmidrule(lr){2-17}
\textbf{Model} & 0 & 1 & 2 & 3 & 4 & 5 & 6 & 7 & 8 & 9 & 10 & 11 & 12 & 13 & 14 & 15 \\
\midrule
\textbf{Base}  & 0.73 & 0.70 & 0.76 & 0.55 & 0.52 & 0.37 & 0.27 & 0.26 & 0.36 & 0.53 & 0.32 & 0.31 & 0.26 & 0.25 & 0.28 & 0.33 \\
\textbf{ReFT}  & 0.73 & 0.83 & 0.72 & 0.57 & 0.66 & 0.50 & 0.36 & 0.37 & 0.25 & 0.50 & 0.47 & 0.41 & 0.42 & 0.35 & 0.46 & 0.51 \\
\textbf{BREP}  & 0.73 & 0.76 & 0.83 & 0.69 & 0.60 & 0.25 & 0.28 & 0.46 & 0.39 & 0.60 & 0.50 & 0.40 & 0.40 & 0.44 & 0.54 & 0.77 \\
\midrule
& \multicolumn{16}{c}{\textbf{Layer}} \\
\cmidrule(lr){2-17}
\textbf{Model} & 16 & 17 & 18 & 19 & 20 & 21 & 22 & 23 & 24 & 25 & 26 & 27 & 28 & 29 & 30 & 31 \\
\midrule
\textbf{Base}  & 0.56 & 0.66 & 0.70 & 0.72 & 0.69 & 0.60 & 0.67 & 0.57 & 0.62 & 0.68 & 0.74 & 0.68 & 0.72 & 0.67 & 0.81 & 0.74 \\
\textbf{ReFT}  & 0.54 & 0.68 & 0.67 & 0.72 & 0.73 & 0.67 & 0.64 & 0.60 & 0.54 & 0.52 & 0.58 & 0.57 & 0.57 & 0.52 & 0.65 & 0.72 \\
\textbf{BREP}  & 0.87 & 0.92 & 0.93 & 0.94 & 0.95 & 0.96 & 0.95 & 0.91 & 0.90 & 0.93 & 0.92 & 0.86 & 0.80 & 0.78 & 0.83 & 0.81 \\
\bottomrule
\end{tabular}
\caption{The detection results of the faithfulness probe at each layer.}
\label{tab:prob_results}
\end{table*}

\begin{table*}[tbp]
\centering
\begin{tabular}{lllllllll}
\toprule

\textbf{Model} & \textbf{Method} & \textbf{BoolQ} & \textbf{PIQA} & \textbf{avg} & \textbf{Physics} & \textbf{Chemistry} & \textbf{Biology} & \textbf{avg} \\
 

\midrule
\multirow{4}{*}{\textbf{Llama3-8B-Instruct}} 
 & \textbf{Base} & 19.4 & 42.4 & 30.9 & 29.1 & 15.1 & \textbf{36.8} & 27.0 \\
  & \textbf{RED} & 15.0\,\textcolor{red}{\scriptsize{↓4.4}} & \textbf{63.9}\,\textcolor{green}{\scriptsize{↑21.5}} & 39.5\,\textcolor{green}{\scriptsize{↑8.6}} & 30.2\,\textcolor{green}{\scriptsize{↑1.1}} & 22.6\,\textcolor{green}{\scriptsize{↑7.5}} & 21.1\,\textcolor{red}{\scriptsize{↓15.7}} & 24.6\,\textcolor{red}{\scriptsize{↓2.4}} \\
 & \textbf{LoRA} & 18.3\,\textcolor{red}{\scriptsize{↓1.1}} & 47.2\,\textcolor{green}{\scriptsize{↑4.8}} & 32.8\,\textcolor{green}{\scriptsize{↑1.9}} & \textbf{34.9}\,\textcolor{green}{\scriptsize{↑5.8}} & 15.1\,\textcolor{blue}{$=$} & 15.8\,\textcolor{red}{\scriptsize{↓21.0}} & 21.9\,\textcolor{red}{\scriptsize{↓5.1}} \\
 & \textbf{BREP} & \textbf{20.7}\,\textcolor{green}{\scriptsize{↑1.3}} & 60.7\,\textcolor{green}{\scriptsize{↑18.3}} & \textbf{40.7}\,\textcolor{green}{\scriptsize{↑9.8}} & 33.7\,\textcolor{green}{\scriptsize{↑4.6}} & \textbf{24.7}\,\textcolor{green}{\scriptsize{↑9.6}} & 26.3\,\textcolor{red}{\scriptsize{↓10.5}} & \textbf{28.2}\,\textcolor{green}{\scriptsize{↑1.2}} \\
\midrule
\multirow{4}{*}{\textbf{Llama-3.1-8B-Instruct}} 
 & \textbf{Base} & 36.4 & 67.9 & 52.2 & 26.7 & 23.7 & 15.8 & 22.1 \\
& \textbf{RED} & 33.8\,\textcolor{red}{\scriptsize{↓2.6}} & 69.4\,\textcolor{green}{\scriptsize{↑1.5}} & 51.6\,\textcolor{red}{\scriptsize{↓0.6}} & 32.6\,\textcolor{green}{\scriptsize{↑5.9}} & 28.0\,\textcolor{green}{\scriptsize{↑4.3}} & \textbf{42.1}\,\textcolor{green}{\scriptsize{↑26.3}} & 34.2\,\textcolor{green}{\scriptsize{↑12.1}} \\
 & \textbf{LoRA} & 35.5\,\textcolor{red}{\scriptsize{↓0.9}} & 69.7\,\textcolor{green}{\scriptsize{↑1.8}} & 52.6\,\textcolor{green}{\scriptsize{↑0.4}} & 33.7\,\textcolor{green}{\scriptsize{↑7.0}} & 31.2\,\textcolor{green}{\scriptsize{↑7.5}} & 36.8\,\textcolor{green}{\scriptsize{↑21.0}} & 33.9\,\textcolor{green}{\scriptsize{↑11.8}} \\
  & \textbf{BREP} & \textbf{41.2}\,\textcolor{green}{\scriptsize{↑4.8}} & \textbf{80.9}\,\textcolor{green}{\scriptsize{↑13.0}} & \textbf{61.1}\,\textcolor{green}{\scriptsize{↑8.9}} & \textbf{40.7}\,\textcolor{green}{\scriptsize{↑14.0}} & \textbf{36.6}\,\textcolor{green}{\scriptsize{↑12.9}} & 26.3\,\textcolor{green}{\scriptsize{↑10.5}} & \textbf{34.5}\,\textcolor{green}{\scriptsize{↑12.4}} \\
\midrule
\multirow{4}{*}{\textbf{Qwen3-8B}} 
 & \textbf{Base} & \textbf{68.9} & 40.1 & \textbf{54.5} & \textbf{58.1} & \textbf{35.5} & 52.6 & \textbf{48.7} \\
  & \textbf{RED} & 70.1\,\textcolor{green}{\scriptsize{↑1.2}} & 23.0\,\textcolor{red}{\scriptsize{↓17.1}} & 46.6\,\textcolor{red}{\scriptsize{↓7.9}} & 60.5\,\textcolor{green}{\scriptsize{↑2.4}} & 31.2\,\textcolor{red}{\scriptsize{↓4.3}} & 57.9\,\textcolor{green}{\scriptsize{↑5.3}} & 49.9\,\textcolor{green}{\scriptsize{↑1.2}} \\
 & \textbf{LoRA} & 69.3\,\textcolor{green}{\scriptsize{↑0.4}} & 40.6\,\textcolor{green}{\scriptsize{↑0.5}} & 55.0\,\textcolor{green}{\scriptsize{↑0.5}} & 46.5\,\textcolor{red}{\scriptsize{↓11.6}} & 30.1\,\textcolor{red}{\scriptsize{↓5.4}} & 68.4\,\textcolor{green}{\scriptsize{↑15.8}} & 48.3\,\textcolor{red}{\scriptsize{↓0.4}} \\
 & \textbf{BREP} & 69.1\,\textcolor{green}{\scriptsize{↑0.2}} & 42.3\,\textcolor{green}{\scriptsize{↑2.2}} & 55.7\,\textcolor{green}{\scriptsize{↑1.2}} & 60.5\,\textcolor{green}{\scriptsize{↑2.4}} & 26.9\,\textcolor{red}{\scriptsize{↓8.6}} & \textbf{63.2}\,\textcolor{green}{\scriptsize{↑10.6}} & 50.2\,\textcolor{green}{\scriptsize{↑1.5}} \\
\midrule
\multirow{4}{*}{\textbf{Qwen3-14B}} 
 & \textbf{Base} & \textbf{71.1} & 49.7 & \textbf{60.4} & \textbf{68.6} & \textbf{35.5} & 52.6 & \textbf{52.2} \\
 & \textbf{RED} & 70.2\,\textcolor{red}{\scriptsize{↓0.9}} & 36.6\,\textcolor{red}{\scriptsize{↓13.1}} & 53.4\,\textcolor{red}{\scriptsize{↓7.0}} & 65.1\,\textcolor{red}{\scriptsize{↓3.5}} & 29.0\,\textcolor{red}{\scriptsize{↓6.5}} & \textbf{68.4}\,\textcolor{green}{\scriptsize{↑15.8}} & 54.2\,\textcolor{green}{\scriptsize{↑2.0}} \\
 & \textbf{LoRA} & \textbf{71.3}\,\textcolor{green}{\scriptsize{↑0.2}} & 51.1\,\textcolor{green}{\scriptsize{↑1.4}} & 61.2\,\textcolor{green}{\scriptsize{↑0.8}} & 58.1\,\textcolor{red}{\scriptsize{↓10.5}} & 26.9\,\textcolor{red}{\scriptsize{↓8.6}} & 57.9\,\textcolor{green}{\scriptsize{↑5.3}} & 47.6\,\textcolor{red}{\scriptsize{↓4.6}} \\
 & \textbf{BREP} & 71.2\,\textcolor{green}{\scriptsize{↑0.1}} & \textbf{53.9}\,\textcolor{green}{\scriptsize{↑4.2}} & \textbf{62.6}\,\textcolor{green}{\scriptsize{↑2.2}} & 62.8\,\textcolor{red}{\scriptsize{↓5.8}} & 36.6\,\textcolor{green}{\scriptsize{↑0.1}} & 57.9\,\textcolor{green}{\scriptsize{↑5.3}} & 
 52.4\,\textcolor{green}{\scriptsize{↑0.2}} \\
\bottomrule
\end{tabular}
\caption{Comparison results of generalization capabilities on commonsense(BoolQ and PIQA) and PIQA(physics, chemistry, and biology). Green arrows \textcolor{green}{(↑)} indicate improvement over base model, red arrows \textcolor{red}{(↓)} indicate decrease. Best method in each group for each metric marked in \textbf{bold}.}
\label{tab:generalization}
\end{table*}

\begin{figure*}[tb]
\centering
\includegraphics[width=1.8\columnwidth]{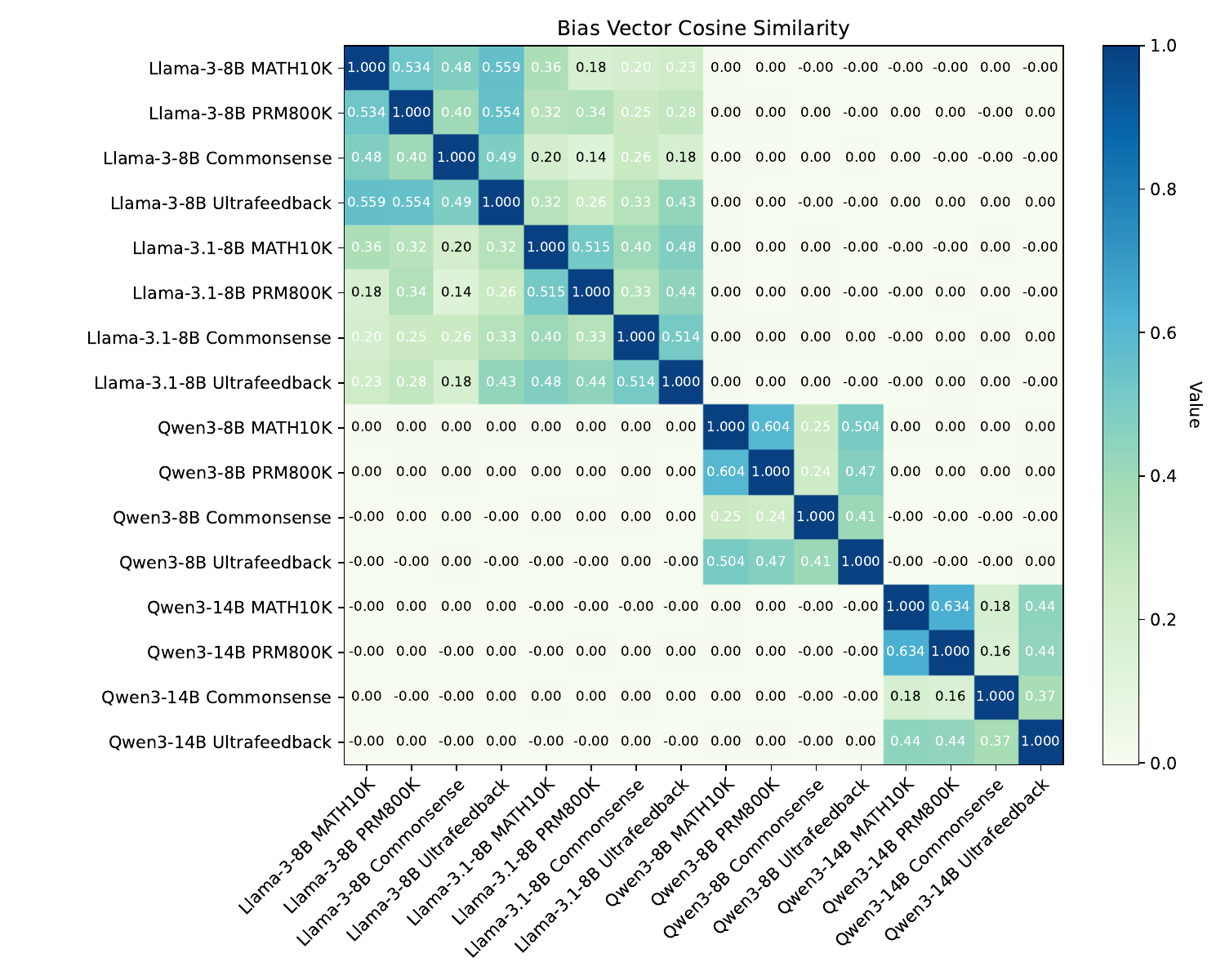} 
\caption{Bias cosine similarity between different models and different tasks.}
\label{cos_map}
\end{figure*}


\end{document}